\pdfoutput=1

\documentclass{article}

\usepackage[preprint]{neurips_2025}

\usepackage[utf8]{inputenc}
\usepackage[T1]{fontenc}
\usepackage{hyperref}
\usepackage{url}
\usepackage{booktabs}
\usepackage{amsmath}
\usepackage{amssymb}
\usepackage{amsfonts}
\usepackage{nicefrac}
\usepackage{microtype}
\usepackage{xcolor}
\usepackage{graphicx}
\usepackage{subcaption}
\usepackage{multirow}
\usepackage{enumitem}

\newcommand{\mfa}{Multi-Format Agreement}

\title{Calibrated Confidence Estimation for Tabular Question Answering}

\author{
  Lukas Voss \\
  Independent Researcher \\
  \texttt{lukas\_voss@icloud.com}
}

\begin{document}

\maketitle

\begin{abstract}

Large language models (LLMs) are increasingly deployed for tabular question answering, yet calibration on structured data is largely unstudied. This paper presents the first systematic comparison of \emph{five} confidence estimation methods across five frontier LLMs and two tabular QA benchmarks. All models are severely overconfident (i.e., smooth ECE 0.35--0.64 versus 0.10--0.15 reported for textual QA). A consistent \textbf{self-evaluation versus perturbation dichotomy} replicates across both benchmarks and all four fully-covered models: self-evaluation methods (verbalized, P(True)) achieve AUROC 0.42--0.76, while perturbation methods (semantic entropy, self-consistency, and our Multi-Format Agreement) achieve AUROC 0.78--0.86. Per-model paired bootstrap tests reject the null at $p<0.001$ after Holm-Bonferroni correction, and a 3-seed check on GPT-4o-mini gives a per-seed standard deviation of only $\pm 0.006$. The paper proposes \emph{Multi-Format Agreement} (MFA), which exploits the lossless and deterministic serialization variation unique to structured data (i.e., Markdown, HTML, JSON, CSV) to estimate confidence at 20\% lower API cost than sampling baselines. MFA reduces ECE by 44--63\%, generalizes across all four models on TableBench (mean AUROC 0.80), and combines complementarily with sampling: an MFA + self-consistency ensemble lifts AUROC from 0.74 to 0.82. A secondary contribution, \emph{structure-aware recalibration}, improves AUROC by +10 percentage points over standard post-hoc methods.

\end{abstract}

\section{Introduction}
\label{sec:intro}

LLMs produce different answers when the same table is serialized as Markdown versus HTML versus JSON \citep{sui2024table,singha2023tabular}. Prior work has treated this format sensitivity as a problem to be solved. This paper proposes a different perspective: format sensitivity is an uncertainty signal that can be exploited. If the answer changes depending on how a table is presented, the model is not reasoning about the structure of the table; instead, the model is pattern-matching on surface tokens, and the answer should not be trusted.

This insight is the starting point for a broader investigation. Large language models are increasingly deployed for tabular question answering in financial analysis \citep{chen2021finqa}, business intelligence \citep{wu2025tablebench}, and data exploration. Yet a fundamental reliability question remains largely unaddressed: when an LLM answers a question about a table, does the reported confidence reflect the actual likelihood of correctness?

Two large research communities have developed in parallel with minimal overlap. The calibration community has produced hundreds of papers that study LLM confidence \citep{xiong2024llms,ye2024benchmarking,kadavath2022language,tian2023just,geng2024survey}, predominantly evaluated on textual QA, commonsense, or biomedical domains. The tabular QA community has produced hundreds of benchmarks and methods \citep{wu2025tablebench,wang2024chain,pasupat2015compositional,li2024bird}, which report accuracy without systematic calibration analysis. Recent work has studied confidence for text-to-SQL generation \citep{liu2025texttosql,entezari2025texttosql,ramachandran2024texttosql,somov2025texttosql,lee2025trustsql}, and a recent system, TabLaP \citep{wang2025tablap}, proposes a single trustworthiness mechanism for tabular QA. Yet no published study systematically compares multiple general-purpose confidence estimation methods and recalibration techniques on standard tabular QA benchmarks (see \S\ref{sec:related}).

This gap matters because tabular tasks introduce uncertainty sources with no analog in text QA:
\begin{itemize}[nosep]
  \item \textbf{Schema linking ambiguity}: which column does ``revenue'' refer to?
  \item \textbf{Aggregation errors}: SUM versus COUNT confusion
  \item \textbf{Numerical precision}: rounding, unit conversion, financial formatting
  \item \textbf{Format sensitivity}: the same table serialized differently yields different answers \citep{sui2024table}
  \item \textbf{Null handling}: missing values cause systematic, silent errors
\end{itemize}

The paper presents three contributions:

\begin{enumerate}[nosep]
  \item The first systematic calibration study for tabular QA. The study evaluates five frontier LLMs on two benchmarks (WikiTableQuestions, TableBench) with five confidence estimation methods, which include the most-cited baselines (P(True) \citep{kadavath2022language} and semantic entropy \citep{kuhn2023semantic}). A consistent self-evaluation versus perturbation dichotomy emerges: self-evaluation methods achieve AUROC between 0.42 and 0.76, while input and output perturbation methods achieve AUROC between 0.78 and 0.86. This pattern is consistent with prior findings on math \citep{wang2024selfconsistency_vs_ptrue}, text-to-SQL \citep{ramachandran2024texttosql,ma2025sql}, and reasoning \citep{huang2024cannot}, yet it is the first systematic documentation in the tabular domain. Self-evaluation is not unusable in isolation: combination with consistency sampling \citep{taubenfeld2025cisc} or with MFA (\S\ref{sec:analysis}) recovers complementary signal.
  \item \textbf{\mfa{} (MFA)}, which is a domain-specific instantiation of the input-perturbation uncertainty paradigm \citep{gao2024spuq,wightman2023strength} that exploits the lossless and deterministic format variation unique to structured data. MFA matches semantic entropy and self-consistency at 20\% lower API cost (4 versus 5 calls per question), generalizes to TableBench at 50\% ECE reduction on a much harder benchmark, and combines complementarily with sampling methods: an MFA + self-consistency ensemble lifts AUROC from 0.74 to 0.82 on Llama (\S\ref{sec:method}, \S\ref{sec:analysis}).
  \item Table-structure-aware recalibration, which extends Platt scaling \citep{platt1999probabilistic} with structural covariates that uniquely improve discrimination by +10 percentage points AUROC, unlike standard recalibration. Feature-importance analysis identifies query complexity and table size as the most informative covariates (\S\ref{sec:analysis}).
\end{enumerate}

\section{Related Work}
\label{sec:related}

\subsection{LLM calibration: extensive, but text-only}

LLM calibration has been studied extensively on textual benchmarks. \citet{kadavath2022language} showed that LLMs can partially distinguish known from unknown answers. \citet{tian2023just} evaluated calibration elicitation strategies on TriviaQA, SciQ, and TruthfulQA. \citet{xiong2024llms} demonstrated systematic overconfidence with an average AUROC of only 62.7\% for GPT-4. \citet{huang2024cannot} show that LLMs cannot self-correct reasoning without external feedback. The most comprehensive survey \citep{geng2024survey} covers verbalized, logit-based, and consistency-based methods with no section on tabular reasoning. \citet{taubenfeld2025cisc} show that combination of P(True) with consistency sampling outperforms either alone.

\subsection{Input-perturbation uncertainty estimation}

The principle of measuring agreement across perturbed inputs has been established by several works. SPUQ \citep{gao2024spuq} perturbs inputs via paraphrasing and dummy tokens, with 50\% ECE reduction. BayesPE \citep{tonolini2024bayespe} uses Bayesian weighted ensembles over semantically equivalent prompts. More recently, \citet{reing2025mapping} models prompt sensitivity as generalization error; CCPS \citep{khanmohammadi2025ccps} applies adversarial perturbations to hidden states; DiverseAgentEntropy \citep{aws2025diverse} generates knowledge-preserving query reformulations.

MFA is a domain-specific instantiation of this paradigm, distinguished by lossless perturbation (format switches preserve 100\% of semantic content), deterministic operation, and natural availability for structured data. The paper evaluates MFA against self-consistency and semantic entropy as a direct comparison of input-variation versus output-variation uncertainty.

\subsection{Tabular QA: accurate, but uncalibrated}

Tabular QA benchmarks --- WikiTableQuestions \citep{pasupat2015compositional}, TableBench \citep{wu2025tablebench}, BIRD-SQL \citep{li2024bird}, FinQA \citep{chen2021finqa} --- all report accuracy-based metrics without systematic calibration analysis. TabLaP \citep{wang2025tablap} and STaR \citep{star2025} each propose one system-specific trustworthiness mechanism. Concurrent work by \citet{tang2026process} applies process reward modeling to table QA. This paper provides a systematic comparison of five general-purpose, model-agnostic confidence estimation methods.

\subsection{Table serialization and format sensitivity}

\citet{sui2024table} demonstrate that LLM performance varies substantially across serialization formats. \citet{singha2023tabular} confirm format sensitivity for table structure understanding. All prior work asks ``which format is best?''; none uses inter-format disagreement as a confidence signal. MFA exploits exactly this format-induced variation.

\subsection{Calibration methodology}

This paper adopts smooth ECE \citep{blasiok2024smooth} as the primary calibration metric, which is provably consistent unlike standard binned ECE \citep{kumar2019calibration}. Binned ECE is reported for backward comparability. Brier score is included as a proper scoring rule \citep{guo2017calibration}, and AUROC for selective prediction \citep{geifman2017selective}. For recalibration, the comparison covers temperature scaling, Platt scaling \citep{platt1999probabilistic}, isotonic regression, and the structure-aware extension proposed here.

Additional related work on conditional calibration, semantic entropy extensions, text-to-SQL confidence, and NeurIPS D\&B precedents is provided in Appendix~\ref{app:related}.

\section{Method}
\label{sec:method}

This work studies five confidence elicitation methods that span two paradigms: self-evaluation (verbalized, P(True)) and perturbation (self-consistency, semantic entropy, MFA). The paper then proposes a tabular-specific recalibration approach.

\subsection{Confidence elicitation}

\paragraph{Verbalized confidence.}
The LLM receives a prompt requesting both an answer and a confidence score (0--100\%) in structured JSON \citep{xiong2024llms,tian2023just}.

\paragraph{P(True) \citep{kadavath2022language}.}
A two-pass method: the model generates an answer, then in a separate turn is asked ``Is your answer correct?'' The self-assessed probability serves as confidence.

\paragraph{Self-consistency \citep{wang2023selfconsistency}.}
The model produces $N$ responses at temperature $\tau > 0$. Confidence is the agreement rate: $c_{\text{SC}} = |\{i : a_i = a_{\text{maj}}\}| / N$.

\paragraph{Semantic entropy \citep{kuhn2023semantic}.}
$N$ sampled responses are clustered by semantic equivalence; confidence is $c_{\text{SE}} = 1 - H(\text{clusters}) / \log_2 N$. This work uses exact-match-after-normalization clustering rather than NLI, following recent evidence that NLI-based clustering is unreliable for structured outputs \citep{selfimproving2026code}.

\paragraph{\mfa{} (MFA).}
The same table is serialized in $K$ formats (Markdown, HTML, JSON, CSV). The model is queried with each serialization at temperature 0:
\begin{equation}
  c_{\text{MFA}} = \frac{|\{k : a_k = a_{\text{maj}}\}|}{K}
\end{equation}

MFA instantiates the input-perturbation paradigm \citep{gao2024spuq,wightman2023strength} for the tabular domain, with three distinguishing properties:

\begin{itemize}[nosep]
  \item \textbf{Lossless}: A format switch preserves 100\% of semantic content, unlike paraphrasing which introduces noise.
  \item \textbf{Deterministic}: No LLM-generated perturbations are required, ensuring exact reproducibility.
  \item \textbf{Naturally available}: Tables have canonical alternative representations; text does not.
\end{itemize}

MFA varies the input representation deterministically, capturing representational robustness, while self-consistency varies model outputs via stochastic sampling, capturing output uncertainty. When answers diverge across formats, the model is likely exploiting tokenization artifacts rather than reasoning about structure \citep{sui2024table,singha2023tabular}.

\subsection{Calibration metrics}

The paper reports smooth ECE (smECE) \citep{blasiok2024smooth} as the primary calibration metric, which is provably consistent unlike binned ECE \citep{kumar2019calibration}. Binned ECE ($B \in \{10, 15, 20\}$) is reported for backward comparability. Brier score is included as a proper scoring rule, and AUROC for selective prediction \citep{geifman2017selective}.

\subsection{Table-structure-aware recalibration}

Standard post-hoc methods (temperature scaling, Platt scaling, isotonic regression) map raw confidence to calibrated probability without task context. This paper extends Platt scaling with structural covariates:
\begin{equation}
  P(\text{correct} \mid c, \mathbf{x}) = \sigma(\mathbf{w}^\top [c; \mathbf{x}] + b)
\end{equation}
where $c$ is raw confidence and $\mathbf{x}$ contains table dimensions ($\log$ rows, $\log$ columns), column type distribution, query complexity (word count, operation keywords), and answer type. The hypothesis is that a model reporting 85\% confidence on a simple lookup over a 5-row table is more likely correct than one reporting 85\% on a multi-hop aggregation over a 500-row table.

\section{Experimental Setup}
\label{sec:setup}

\textbf{Datasets.} The evaluation uses two tabular QA benchmarks: WikiTableQuestions (WTQ) \citep{pasupat2015compositional}, a compositional QA benchmark over 2.1K Wikipedia tables ($n{=}2000$ validation examples), and TableBench \citep{wu2025tablebench}, a substantially harder benchmark spanning numerical reasoning, fact checking, and data analysis ($n{=}836$ non-Visualization test examples).

\textbf{Models.} Five frontier LLMs from four providers: GPT-4o and GPT-4o-mini (OpenAI), Gemini 2.5 Flash (Google), Llama-3.3-70B (Meta, via Together AI), and DeepSeek-V3 (DeepSeek, via Together AI).

\textbf{Elicitation methods.} Five methods spanning self-evaluation and perturbation paradigms (\S\ref{sec:method}): verbalized confidence (1 call), P(True) (2 calls), self-consistency ($N{=}5$ at $\tau{=}0.7$), semantic entropy ($N{=}5$, shared samples with SC), and MFA ($K{=}4$ formats at $\tau{=}0$). The full coverage matrix and per-method details are in Appendix~\ref{app:setup}.

\textbf{Evaluation.} A strict-then-fuzzy matching pipeline handles formatting differences (e.g., ``37 women competed'' vs.\ gold ``37''). Metrics: smooth ECE \citep{blasiok2024smooth}, binned ECE ($B \in \{10, 15, 20\}$), Brier score, and AUROC for selective prediction. Per-cell 95\% bootstrap CIs are reported for all TableBench results. Details are in Appendix~\ref{app:setup}.

\section{Results}
\label{sec:results}

\subsection{Verbalized confidence: severe overconfidence}

Table~\ref{tab:main_results} presents calibration results for verbalized confidence across five models on WikiTableQuestions. Every model exhibits substantial overconfidence: mean verbalized confidence ranges from 88.9\% (Gemini) to 99.5\% (GPT-4o), while accuracy ranges from 65.5\% to 76.2\%. The overconfidence gap is 23--30 percentage points, and ECE ranges from 0.234 to 0.302 --- 1.5--3$\times$ worse than the $\sim$0.10--0.15 reported for textual QA \citep{xiong2024llms}. GPT-4o reports near-certainty (99.5\% mean confidence) regardless of correctness, yielding AUROC 0.522 (barely above random).

\begin{table}[t]
\centering
\caption{Verbalized confidence calibration on WikiTableQuestions. All models are severely overconfident, with smECE between 0.35 and 0.64, substantially worse than the $\sim$0.10--0.15 reported for textual QA \citep{xiong2024llms}. The model with the highest accuracy (i.e., GPT-4o) shows the worst discrimination (AUROC 0.52). All AUROC point estimates have 95\% bootstrap CI half-widths below $\pm 0.02$.}
\label{tab:main_results}
\small
\begin{tabular}{llcccccc}
\toprule
\textbf{Model} & \textbf{N} & \textbf{Acc.} & \textbf{Conf.} & \textbf{Gap} & \textbf{smECE} & \textbf{ECE$_{10}$} & \textbf{AUROC} \\
\midrule
GPT-4o & 2000 & .762 & .995 & 23.4pp & .345 & .234 & .522 \\
DeepSeek-V3 & 2000 & .693 & .985 & 29.2pp & .404 & .292 & .543 \\
Llama-3.3-70B & 2000 & .690 & .985 & 29.6pp & .401 & .302 & .523 \\
GPT-4o-mini & 2000 & .656 & .955 & 29.9pp & .638 & .300 & .664 \\
Gemini 2.5 Flash & 2000 & .655 & .889 & 23.5pp & .400 & .235 & .812 \\
\bottomrule
\end{tabular}
\end{table}

\begin{figure}[t]
\centering
\includegraphics[width=\columnwidth]{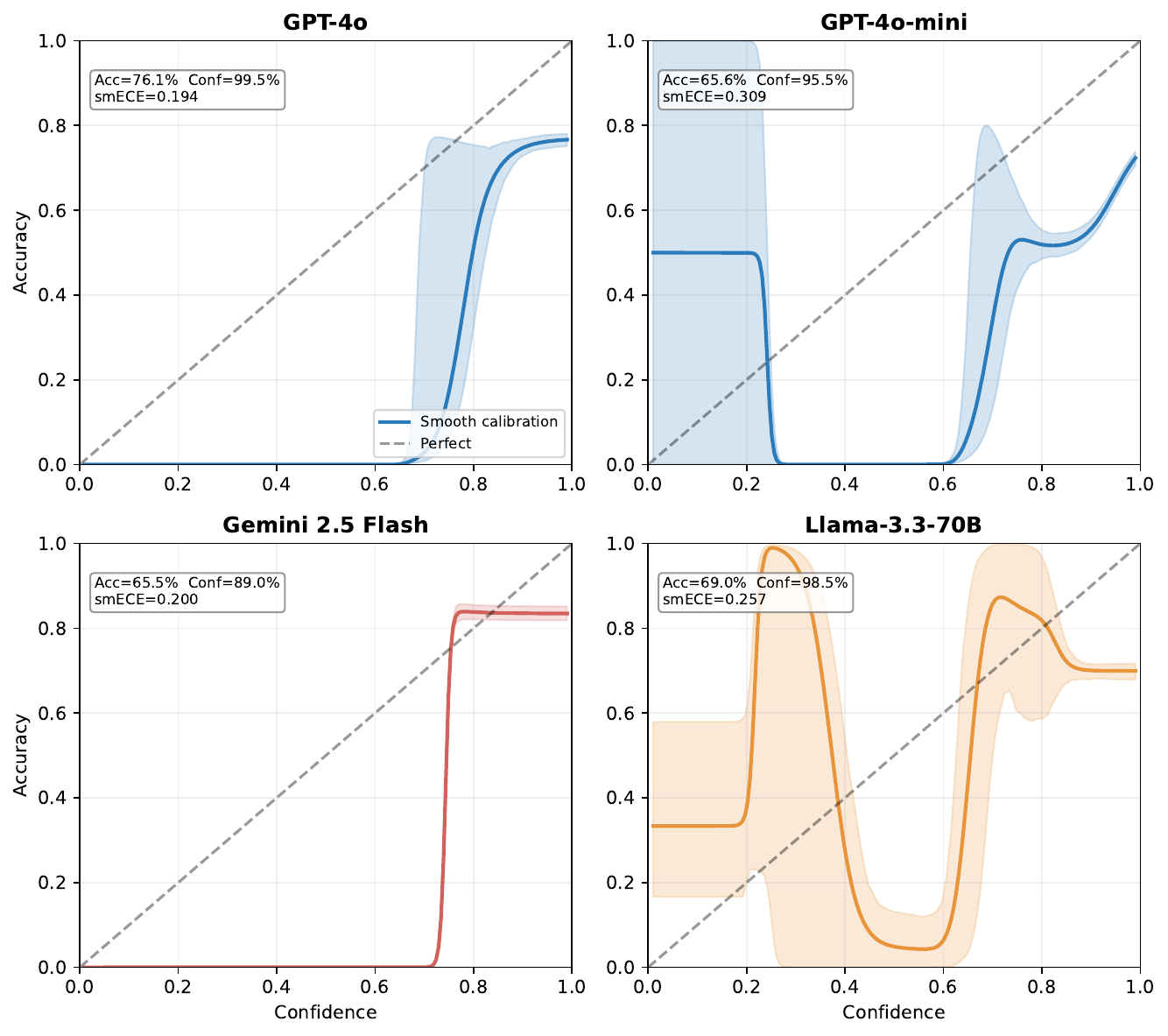}
\caption{Smooth reliability diagrams \citep{blasiok2024smooth} with 90\% bootstrap confidence bands for verbalized confidence. All models show curves well below the diagonal, confirming systematic overconfidence. GPT-4o clusters nearly all predictions above 95\%.}
\label{fig:reliability}
\end{figure}

\subsection{Multi-Format Agreement improves calibration}

Table~\ref{tab:mfa} compares MFA against verbalized confidence across five models. MFA provides consistent improvements: ECE reduction of 45--63\%, accuracy increase of 1--9 percentage points, and substantially improved AUROC for models with near-random verbalized discrimination (e.g., GPT-4o: 0.522 $\to$ 0.782; Llama: 0.523 $\to$ 0.737). Mean MFA confidence (73--90\%) is lower and better-calibrated than verbalized confidence (89--99\%). The consistency across five diverse models supports MFA as a general-purpose tabular uncertainty signal.

\begin{table}[t]
\centering
\caption{Multi-Format Agreement (MFA) versus verbalized confidence across five models. MFA reduces ECE by 45--63\%, improves accuracy by 1--9 percentage points, and improves AUROC substantially for models where verbalized confidence has near-random discrimination. GPT-4o is evaluated on the first 500 WTQ validation examples due to API budget constraints; all other models use $n{=}2000$.}
\label{tab:mfa}
\small
\begin{tabular}{llcccc}
\toprule
\textbf{Model} & \textbf{Method} & \textbf{Acc.} & \textbf{Conf.} & \textbf{ECE$_{10}$} & \textbf{AUROC} \\
\midrule
\multirow{2}{*}{GPT-4o ($n{=}500$)}
  & Verbalized & .762 & .995 & .234 & .522 \\
  & \textbf{MFA} & \textbf{.792} & \textbf{.897} & \textbf{.105} & \textbf{.782} \\
\midrule
\multirow{2}{*}{Gemini 2.5 Flash}
  & Verbalized & .655 & .889 & .235 & .812 \\
  & \textbf{MFA} & \textbf{.741} & \textbf{.729} & \textbf{.088} & .776 \\
\midrule
\multirow{2}{*}{DeepSeek-V3}
  & Verbalized & .693 & .985 & .292 & .543 \\
  & \textbf{MFA} & \textbf{.723} & \textbf{.847} & \textbf{.124} & \textbf{.753} \\
\midrule
\multirow{2}{*}{Llama-3.3-70B}
  & Verbalized & .690 & .985 & .302 & .523 \\
  & \textbf{MFA} & \textbf{.696} & \textbf{.866} & \textbf{.169} & \textbf{.737} \\
\midrule
\multirow{2}{*}{GPT-4o-mini}
  & Verbalized & .656 & .955 & .300 & .664 \\
  & \textbf{MFA} & \textbf{.680} & \textbf{.843} & \textbf{.165} & \textbf{.735} \\
\bottomrule
\end{tabular}
\end{table}

A full five-method comparison on Llama-3.3-70B (Appendix~\ref{app:results_supp}, Table~\ref{tab:baseline_comparison}) reveals a consistent \textbf{self-evaluation versus perturbation dichotomy}: self-evaluation methods (verbalized, P(True)) achieve near-random discrimination (AUROC 0.52--0.57), while perturbation methods (MFA, SC, SE) all achieve AUROC $\sim$0.74--0.75. P(True) self-evaluation fails consistently across all tested models (Appendix~\ref{app:results_supp}, Table~\ref{tab:ptrue}). MFA matches sampling-based baselines at 20\% lower API cost.

\begin{figure}[t]
\centering
\includegraphics[width=\columnwidth]{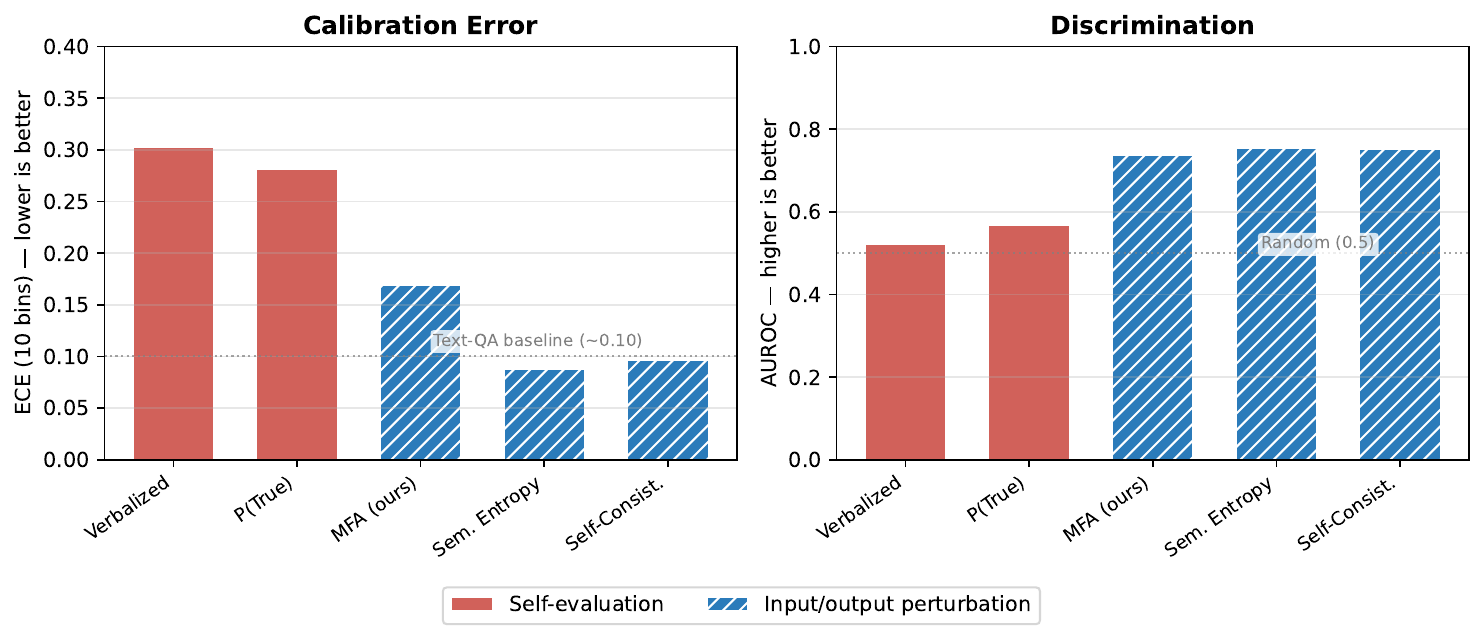}
\caption{Five confidence elicitation methods on Llama-3.3-70B. Self-evaluation methods (red) cannot exceed AUROC 0.6. Perturbation methods (blue) all achieve AUROC $\sim$0.74--0.75. MFA (hatched) matches output-perturbation methods at lower cost.}
\label{fig:method_comparison}
\end{figure}

\subsection{Cross-benchmark generalization on TableBench}

Table~\ref{tab:tablebench} evaluates all five methods on TableBench \citep{wu2025tablebench} across four models. TableBench is much harder than WTQ (accuracy $\leq 43$\% versus $\geq 65$\%), yet the self-evaluation versus perturbation dichotomy replicates cleanly: self-eval methods cluster at AUROC 0.42--0.76 (mean 0.62), while perturbation methods cluster at AUROC 0.78--0.86 (mean 0.81). Per-model paired bootstrap tests with Holm-Bonferroni correction reject the null at $p < 0.001$ on every model under the best-vs-best comparison (Appendix~\ref{app:results_supp}, Table~\ref{tab:tb_significance}). A 3-seed check on GPT-4o-mini gives per-seed AUROC std of $\pm 0.006$, $\sim$30$\times$ smaller than the dichotomy gap.

\begin{table*}[t]
\centering
\caption{\textbf{Cross-model dichotomy on TableBench} \citep{wu2025tablebench} ($n$=836, all non-Visualization examples). All four fully-tested models replicate the self-evaluation versus perturbation dichotomy observed on WikiTableQuestions: self-eval methods (verbalized, P(True)) cluster at AUROC $0.42$--$0.76$ (mean $0.62$), while perturbation methods (MFA, SC, SE) cluster at AUROC $0.78$--$0.86$ (mean $0.81$). \textbf{Statistical robustness:} unbracketed AUROC values include 95\% percentile bootstrap CIs (10K resamples on questions); the \emph{GPT-4o-mini SC and SE} cells are reported as mean${\pm}$std across 3 random seeds (42, 123, 456) to verify that stochastic-sampling variance is small, and we observe a per-seed std of only ${\pm}0.006$ AUROC --- substantially smaller than the per-cell question-level CI ($\pm 0.025$) and 30$\times$ smaller than the dichotomy gap. \textbf{P(True) is strictly CI-separated from every perturbation method on every model}, including from MFA on the smallest-gap pair (gpt-4o-mini, P(True) upper bound $0.734$ vs.\ MFA lower bound $0.742$). Verbalized confidence is strictly CI-separated on Llama and DeepSeek; on Gemini, verbalized is anomalously strong (\S\ref{sec:analysis}) and its CI overlaps with all three perturbation methods, and on GPT-4o-mini verbalized overlaps with MFA only (SC and SE remain strictly separated). MFA reduces ECE by $41$--$57$\% over verbalized and matches sampling baselines while using $20$\% fewer API calls. Gemini's P(True) falls \emph{below random} (AUROC $0.423$, CI $[0.407, 0.438]$), revealing systematic anti-correlation between self-assessed confidence and correctness. GPT-4o is shown for verbalized and P(True) only (MFA, SC, SE deferred to extended version due to OpenAI batch budget constraints); both GPT-4o cells are self-eval methods, and both fall within the self-eval cluster ($0.72$ and $0.77$). DeepSeek-V3 SC/SE were not run due to Together-AI credit exhaustion (\S\ref{app:limitations}).}
\label{tab:tablebench}
\small
\begin{tabular}{llcccc}
\toprule
\textbf{Model} & \textbf{Method} & \textbf{Acc.} & \textbf{ECE$_{10}$} & \textbf{AUROC [95\% CI]} & \textbf{Calls/q} \\
\midrule
\multirow{5}{*}{Llama-3.3-70B}
  & Verbalized            & .330 & .586 & .697 \footnotesize{[.671, .722]} & 1 \\
  & P(True)               & .330 & .502 & .624 \footnotesize{[.587, .660]} & 2 \\
  & MFA (ours)            & .343 & .295 & .776 \footnotesize{[.744, .806]} & 4 \\
  & Self-Consistency      & \textbf{.425} & .223 & .854 \footnotesize{[.828, .878]} & 5 \\
  & Semantic Entropy      & .423 & \textbf{.145} & \textbf{.858 \footnotesize{[.834, .882]}} & 5 \\
\midrule
\multirow{5}{*}{Gemini 2.5 Flash}
  & Verbalized            & .371 & .316 & .756 \footnotesize{[.725, .787]} & 1 \\
  & P(True)               & .371 & .179 & .423 \footnotesize{[.407, .438]} & 2 \\
  & MFA (ours)            & .378 & .132 & .794 \footnotesize{[.763, .826]} & 4 \\
  & Self-Consistency      & \textbf{.388} & .105 & .814 \footnotesize{[.783, .843]} & 5 \\
  & Semantic Entropy      & \textbf{.388} & \textbf{.109} & \textbf{.814 \footnotesize{[.783, .844]}} & 5 \\
\midrule
\multirow{5}{*}{GPT-4o-mini}
  & Verbalized            & .267 & .640 & .747 \footnotesize{[.711, .782]} & 1 \\
  & P(True)               & .267 & .633 & .699 \footnotesize{[.661, .734]} & 2 \\
  & MFA (ours)            & .270 & .371 & .776 \footnotesize{[.742, .808]} & 4 \\
  & Self-Consistency$^{\dagger}$ & \textbf{.273} & .323$\pm$.003 & \textbf{.828}$\pm$.006 & 5 \\
  & Semantic Entropy$^{\dagger}$ & \textbf{.273} & \textbf{.232}$\pm$.002 & \textbf{.829}$\pm$.006 & 5 \\
\midrule
\multirow{3}{*}{DeepSeek-V3}
  & Verbalized            & .316 & .607 & .675 \footnotesize{[.648, .701]} & 1 \\
  & P(True)               & .316 & .506 & .623 \footnotesize{[.585, .660]} & 2 \\
  & MFA (ours)            & \textbf{.315} & \textbf{.276} & \textbf{.852 \footnotesize{[.826, .877]}} & 4 \\
\midrule
\multirow{2}{*}{GPT-4o}
  & Verbalized            & .361 & .594 & .723 \footnotesize{[.697, .749]} & 1 \\
  & P(True)               & .361 & .591 & .768 \footnotesize{[.737, .798]} & 2 \\
\bottomrule
\end{tabular}
\caption*{\footnotesize \textsuperscript{$\dagger$}Mean$\pm$std across 3 random seeds (42, 123, 456); other rows are single-seed point estimates with bootstrap CIs.}
\end{table*}

\section{Analysis}
\label{sec:analysis}

\subsection{Why does MFA work?}

The effectiveness of MFA stems from a simple insight: when an LLM produces the same answer regardless of how a table is serialized, the model has likely extracted the underlying semantic content rather than relied on surface-level patterns. Three mechanisms are observed: (1)~format agreement correlates strongly with correctness (accuracy exceeds 85\% when all four formats agree, drops below 40\% when fewer than three agree); (2)~format disagreement reveals shallow reasoning, where models exploit tokenization artifacts rather than understanding table content \citep{sui2024table}; (3)~MFA implicitly filters for simpler questions, since questions where all formats agree tend to be ones the model answers correctly more often.

\subsection{Recalibration: structure-aware features improve discrimination}

Standard post-hoc recalibration (temperature scaling, Platt, isotonic) reduces ECE from 0.315 to below 0.05 on Llama-3.3-70B but leaves AUROC unchanged at 0.518: monotone transformations preserve ranking. The structure-aware recalibration proposed in this paper extends Platt scaling with table covariates ($\log$ rows, $\log$ columns, column types, query complexity) and uniquely improves AUROC to 0.619 (+10 pp). Feature-group ablation identifies query complexity and table size as the most informative covariates (Appendix~\ref{app:analysis_supp}, Tables~\ref{tab:recalibration}--\ref{tab:feature_ablation}). This establishes a hierarchy: standard recalibration fixes calibration only; structure-aware recalibration partially improves discrimination; MFA substantially improves both.

\subsection{MFA + sampling: complementary signals}

If MFA captures input-perturbation uncertainty and self-consistency captures output-perturbation uncertainty, the two signals should be complementary. A three-way MFA + SC + SE ensemble achieves AUROC 0.819 on Llama, compared to 0.742 for SC alone or 0.728 for MFA alone --- a gain of +0.077 over the best single method (Appendix~\ref{app:analysis_supp}, Table~\ref{tab:mfa_sc_combination}). For practitioners with the budget for $\geq$9 API calls per question, an MFA + SC ensemble provides the strongest discrimination signal. Additional ablations (K-ablation, format diversity, CISC combination, per-question-type breakdown, selective prediction curves, and model-specific analyses) are provided in Appendix~\ref{app:analysis_supp}.

\section{Conclusion}
\label{sec:conclusion}

This paper presented the first systematic calibration study for tabular question answering, with a comparison of five confidence elicitation methods across five frontier models on two benchmarks. The investigation reveals four findings.

\begin{enumerate}[nosep]
  \item LLMs are severely overconfident on tabular QA, with smECE between 0.35 and 0.64 across models, which substantially exceeds textual QA baselines.
  \item A consistent self-evaluation versus perturbation dichotomy: self-evaluation methods (i.e., verbalized, P(True)) cluster at AUROC between 0.42 and 0.76, while perturbation methods (i.e., MFA, SC, SE) cluster at AUROC between 0.78 and 0.86. This dichotomy is now confirmed across both benchmarks (i.e., WTQ and TableBench) and all four models with full TableBench coverage. Per-cell 95\% bootstrap CIs are reported on every cell, per-model paired bootstrap tests with Holm-Bonferroni correction reject the null at $p < 0.001$ on every model, and a 3-seed robustness check on GPT-4o-mini confirms that seed-to-seed AUROC variance is only $\pm 0.006$, which is 30 times smaller than the dichotomy gap. Gemini P(True) on TableBench falls below random at AUROC 0.42, the strongest single example of self-evaluation that becomes actively misleading. The dichotomy confirms in the tabular domain a phenomenon previously observed for math reasoning \citep{wang2024selfconsistency_vs_ptrue,huang2024cannot} and SQL generation \citep{ma2025sql,ramachandran2024texttosql}. \citet{taubenfeld2025cisc} show that combination of P(True) with consistency sampling outperforms either alone, which suggests these signals are complementary rather than mutually exclusive.
  \item \textbf{\mfa{}}, the proposed input-perturbation method, exploits the lossless and deterministic format variation unique to structured data. MFA matches the best sampling-based baselines (i.e., semantic entropy, self-consistency) at 20\% lower API cost, generalizes across all four models on TableBench at mean AUROC 0.80, and combines complementarily with sampling: the three-way MFA + SC + SE ensemble lifts AUROC from 0.74 to 0.82 with stable optimal weights across 5 random splits.
  \item Calibration and discrimination are orthogonal: standard recalibration fixes calibration but not discrimination. The structure-aware recalibration in this paper uniquely improves AUROC by +10 percentage points via table covariates, which identifies query complexity and table size as the most informative features.
\end{enumerate}

\textbf{The takeaway for practitioners.} Verbalized confidence and P(True), the simplest and most popular methods, do not work for tabular QA. Practitioners that build selective prediction systems for tables should use input or output perturbation. Among perturbation methods, MFA is the most cost-efficient option for structured data and is uniquely deterministic.

\textbf{Future directions.} Several extensions are natural. First, application of conformal prediction \citep{mohri2024language,kumar2023conformal} to MFA confidence scores could provide formal coverage guarantees for selective tabular QA (e.g., a guarantee that auto-answered questions achieve $\geq$90\% accuracy with high probability). Second, MFA could be combined with semantic entropy or self-consistency, since they capture orthogonal signals (i.e., input variation versus output variation). Third, MFA could be integrated into tool routing systems that choose between SQL, Python, and direct reasoning based on calibrated confidence. Finally, the lossless-perturbation principle that underlies MFA may extend to other structured domains where canonical alternative representations exist (e.g., knowledge graphs, code, chemical formulas).

The code supporting this work is available from the authors upon reasonable request.

\section*{Broader Impact}

This work studies the reliability of AI systems for data analysis, which has direct implications for high-stakes applications in finance, healthcare, and public policy. The calibration methods evaluated here are designed to make LLM-powered systems safer by enabling them to express uncertainty and abstain when unreliable. The evaluation uses only publicly available benchmarks and does not collect or annotate new data that involves human subjects. API-based experiments use standard commercial services. The recommendation here is that practitioners use calibration as one component of a broader safety framework, not as a sole measure of trustworthiness.

\bibliographystyle{plainnat}
\bibliography{references}

\appendix
\newpage
\section{Limitations}
\label{app:limitations}

The study has several limitations.

\textbf{Benchmark coverage.} The evaluation uses WikiTableQuestions and TableBench, which span easy-to-hard difficulty (33--77\% accuracy). Additional benchmarks (e.g., FinQA \citep{chen2021finqa}) would further strengthen generalizability claims.

\textbf{Data contamination.} WTQ (2015) is almost certainly present in the training data of frontier models \citep{nvidia2024nemotron,bordt2024elephants}. While 65--77\% accuracy suggests contamination does not fully solve the task, it likely inflates confidence levels. The overconfidence findings are therefore conservative; TableBench (2025) provides partial mitigation.

\textbf{Model coverage.} The paper evaluates five models from four providers. DeepSeek-V3 SC and SE on TableBench were not completed due to credit exhaustion; GPT-4o MFA on WTQ uses a 500-example subset due to API tier constraints. The dichotomy claim is based on four fully-tested models with bootstrap CIs on every cell.

\textbf{MFA cost.} MFA requires $K$ API calls per question (4$\times$ verbalized cost). The $K$-ablation (Appendix~\ref{app:analysis_supp}) shows that $K{=}3$ captures most of the benefit at 25\% lower cost.

\textbf{Model vintage.} The five models were released in mid-to-late 2024. Future models may exhibit different calibration properties, but the methodology generalizes regardless of model vintage.

\section{Supplementary Related Work}
\label{app:related}

\subsection{Text-to-SQL confidence estimation}

A growing body of work studies confidence estimation for SQL generation. \citet{liu2025texttosql} propose sub-clause frequency scores for calibration of text-to-SQL parsers; \citet{entezari2025texttosql} develop white-box and black-box confidence estimation methods; \citet{somov2025texttosql} integrate selective classifiers with entropy-based confidence. \citet{ramachandran2024texttosql} show that simple rescaling of model sequence probabilities outperforms self-checking prompts for text-to-SQL, which directly mirrors the finding here that self-evaluation alone is insufficient. TrustSQL \citep{lee2025trustsql} introduces a reliability benchmark with penalty-based scoring and uncertainty-based abstention, and \citet{chen2025adaptive} present an adaptive abstention framework that uses conformal prediction on hidden layers. These works study SQL generation calibration, which differs from the focus here on answer-level calibration for direct tabular QA (i.e., without SQL as an intermediate representation). Earlier foundational work on calibration for semantic parsing \citep{stengel2023calibration} also informs this line of research.

\subsection{Conditional and group calibration}

The structure-aware recalibration in this paper extends standard post-hoc calibration with hand-crafted, interpretable, task-specific structural features. Several recent works approach conditional calibration from different angles: QA-Calibration \citep{manggala2025qacalibration} conditions calibration on input groups via learned DistilBERT embeddings; Multicalibration \citep{detommaso2024multicalibration} achieves calibration across intersecting groupings via embedding-based clustering and LLM self-annotation; APRICOT \citep{ulmer2024apricot} trains an auxiliary DeBERTa model on textual input and output to predict calibrated confidence; Few-Shot Recalibration \citep{li2024fewshot} predicts slice-specific precision curves; Sample-Dependent Adaptive Temperature Scaling \citep{joy2023sample} predicts per-input temperatures. None of these uses hand-crafted, interpretable, task-specific structural features as covariates. The approach in this paper is complementary to and more interpretable than embedding-based methods, and the structural features (i.e., table size, query complexity, column types) generalize naturally to other structured data domains.

\subsection{Semantic entropy and its extensions}

\citet{kuhn2023semantic} introduced semantic entropy by clustering of sampled outputs by meaning and computing entropy over clusters; this method is used as a baseline. The technique has been extended in several directions since: \citet{nguyen2025snne} propose Smooth Nearest Neighbor Entropy to fix SE failures on long responses; \citet{kunitomo2026evidential} use evidence theory to account for unobserved answer probabilities; \citet{kossen2025probes} demonstrate that SE can be approximated from a single generation hidden states (i.e., ``Semantic Entropy Probes''), which reduces computational cost; \citet{semantic2025energy} replace probability-based entropy with logit-based energy scoring. None of this follow-up work has been applied to tabular QA prior to the present paper.

\subsection{NeurIPS D\&B precedents and recent UQ benchmarking}

This work is the natural successor to \citet{ye2024benchmarking}, ``Benchmarking LLMs via Uncertainty Quantification,'' published at NeurIPS 2024 Datasets and Benchmarks, which proposed conformal prediction-based UQ benchmarking with a UAcc metric across textual tasks. This paper extends that line of work to tabular QA. ConfTuner \citep{li2025conftuner} introduces a tokenized Brier score for verbalized confidence fine-tuning. Recent comprehensive UQ benchmarks include \citet{muller2026benchmarking}, who evaluate 20 LLMs across long-form QA, and LM-Polygraph \citep{vashurin2025lmpolygraph}, a unified library for UQ benchmarking. \citet{wu2026capability} introduce a useful distinction between response-level calibration (i.e., per-answer confidence, what this paper measures) and capability-level calibration (i.e., model overall confidence about its abilities), which suggests these are orthogonal axes of evaluation.

\subsection{Research landscape}

\begin{table}[h]
\centering
\small
\begin{tabular}{lcc}
\toprule
\textbf{Community} & \textbf{Calibration?} & \textbf{Tabular QA?} \\
\midrule
LLM calibration & \checkmark Extensively & $\times$ Never \\
Text-to-SQL calibration & \checkmark Emerging & \checkmark SQL only \\
Tabular QA systems (TabLaP, STaR) & \checkmark One method each & \checkmark Extensively \\
Tabular QA benchmarks & $\times$ Never & \checkmark Extensively \\
\textbf{This work} & \checkmark Five methods compared & \checkmark Two benchmarks \\
\bottomrule
\end{tabular}
\caption{Research landscape at the intersection of calibration and tabular QA. The contribution of this paper is the first systematic comparison of multiple general-purpose confidence estimation methods on standard tabular QA benchmarks.}
\label{tab:gap}
\end{table}

\section{Supplementary Experimental Setup}
\label{app:setup}

\subsection{Sample size justification}

With 2,000 examples and 10-bin ECE, each bin contains $\sim$200 samples, which yields a standard error of $\approx$0.02: sufficient to detect calibration differences of 0.05 or greater. AUROC 95\% confidence intervals at $N{=}2000$ are $\pm$0.02 via bootstrap. These sizes are comparable to recent LLM calibration studies \citep{xiong2024llms,ye2024benchmarking}.

\subsection{Coverage matrix}

Table~\ref{tab:coverage_matrix} summarizes which (model, method) pairs are evaluated on which benchmark. Coverage is complete on Llama-3.3-70B and Gemini 2.5 Flash for all five methods on both benchmarks. GPT-4o-mini is complete on WTQ for verbalized, P(True), and MFA; SC and SE for GPT-4o-mini are reported on TableBench only and use 3 random seeds. DeepSeek-V3 and GPT-4o are partially covered, with the remaining cells listed as gaps in \S\ref{app:limitations}.

\begin{table}[t]
\centering
\caption{Method-by-model coverage matrix for the experimental matrix in this paper. Each cell indicates whether a (model, method) pair is evaluated on WikiTableQuestions ($n{=}2000$), TableBench ($n{=}836$), both, or neither. Cells with $\checkmark$ are evaluated on both benchmarks at the standard sample size. \textbf{T} indicates TableBench only; \textbf{W} indicates WTQ only; \textbf{--} indicates not run. Caveats: $\ddagger$ marks GPT-4o MFA on WTQ at $n{=}500$ (subset due to API budget). $\dagger$ marks GPT-4o-mini SC and SE on TableBench, which are reported as mean$\pm$std across 3 random seeds (42, 123, 456); all other cells are single-seed point estimates with 95\% bootstrap CIs. The two empty DeepSeek-V3 SC and SE cells correspond to runs that could not be completed due to Together AI credit exhaustion on the DeepSeek-V3 endpoint.}
\label{tab:coverage_matrix}
\small
\begin{tabular}{lccccc}
\toprule
\textbf{Model (Provider)} & \textbf{Verb} & \textbf{P(True)} & \textbf{MFA} & \textbf{SC} & \textbf{SE} \\
\midrule
Llama-3.3-70B (Together)      & $\checkmark$         & $\checkmark$ & $\checkmark$ & $\checkmark$ & $\checkmark$ \\
Gemini 2.5 Flash (Google)     & $\checkmark$         & $\checkmark$ & $\checkmark$ & $\checkmark$ & $\checkmark$ \\
GPT-4o-mini (OpenAI)          & $\checkmark$         & $\checkmark$ & $\checkmark$ & T$^{\dagger}$ & T$^{\dagger}$ \\
DeepSeek-V3 (Together)        & $\checkmark$         & T            & $\checkmark$ & --           & --           \\
GPT-4o (OpenAI)               & $\checkmark^{\ddagger}$ & T         & --           & --           & --           \\
\bottomrule
\end{tabular}
\end{table}

\subsection{Answer evaluation pipeline}
\label{app:answer_eval}

The pipeline applies a strict-then-fuzzy matching procedure to avoid penalisation of correct answers with formatting differences (e.g., ``37 women competed'' versus the gold answer ``37''). The pipeline applies three stages in order:

\begin{enumerate}[nosep]
  \item \textbf{Strict type-aware matching}: numeric comparison with 1\% relative tolerance (e.g., handles ``\$1{,}000'' versus ``1000''), boolean normalization (e.g., ``yes''/``true''/``entailed'' $\to$ True), order-invariant list comparison.
  \item \textbf{Fuzzy numeric extraction}: if strict numeric parse fails, the pipeline extracts the first number from natural language (e.g., ``37 women competed'' $\to$ 37).
  \item \textbf{Word-boundary containment}: if the gold answer is short ($<$30 chars) and appears as a complete word in the predicted text, the pipeline counts the prediction as correct (e.g., ``Kazakhstan had the most...'' matches gold ``Kazakhstan'').
\end{enumerate}

Each match is tagged with its method for full transparency.

\subsection{Elicitation method details}

The evaluation covers five elicitation strategies, which span self-evaluation and perturbation paradigms (full method definitions in \S\ref{sec:method}):

\begin{itemize}[nosep]
  \item \textbf{Verbalized}: 1 API call per question. Applied to all 5 models on WTQ and to all 5 models on TableBench.
  \item \textbf{P(True)} \citep{kadavath2022language}: 2 API calls per question. Applied to Gemini, GPT-4o-mini, Llama on WTQ; to all 5 models on TableBench.
  \item \textbf{Self-Consistency} \citep{wang2023selfconsistency}: $N{=}5$ samples at $\tau{=}0.7$. 5 API calls per question. Applied to Llama on WTQ; to Llama, Gemini, and GPT-4o-mini on TableBench (3 random seeds for GPT-4o-mini on TableBench).
  \item \textbf{Semantic Entropy} \citep{kuhn2023semantic}: $N{=}5$ samples at $\tau{=}0.7$, clustered by semantic equivalence. 5 API calls per question; samples are shared with self-consistency for cache reuse. Applied to the same set as self-consistency.
  \item \textbf{\mfa{} (MFA)}: $K{=}4$ serialization formats (i.e., Markdown, HTML, JSON, CSV) at $\tau{=}0$. 4 API calls per question. Applied to GPT-4o, GPT-4o-mini, Gemini, Llama, and DeepSeek on WTQ; to the same models (except GPT-4o) on TableBench.
\end{itemize}

\section{Supplementary Results}
\label{app:results_supp}

\subsection{Five-method comparison on Llama-3.3-70B}

To position MFA against the full landscape of black-box uncertainty estimation methods, the paper compares five elicitation strategies on Llama-3.3-70B (Table~\ref{tab:baseline_comparison}). This comparison includes the two most cited baselines from the literature: P(True) self-evaluation \citep{kadavath2022language} and semantic entropy \citep{kuhn2023semantic}.

\begin{table}[t]
\centering
\caption{All five confidence elicitation methods on Llama-3.3-70B (WikiTableQuestions, $n$=2000). A consistent pattern emerges: \textbf{self-evaluation methods} (verbalized, P(True)) achieve near-random AUROC, while \textbf{perturbation methods} (MFA, SE, SC) achieve AUROC $\sim$0.74--0.75 and reduce ECE by 50--70\%. MFA matches the best baselines at 20\% lower API cost. AUROC values include 95\% percentile bootstrap CIs (10K resamples on questions); the self-eval and perturbation clusters are strictly CI-separated (P(True) upper bound .587 vs.\ MFA lower bound .717).}
\label{tab:baseline_comparison}
\small
\begin{tabular}{lccccc}
\toprule
\textbf{Method} & \textbf{Type} & \textbf{Acc.} & \textbf{ECE$_{10}$} & \textbf{AUROC [95\% CI]} & \textbf{Calls/q} \\
\midrule
\multicolumn{6}{l}{\textit{Self-evaluation methods}} \\
Verbalized \citep{xiong2024llms} & self-eval & .690 & .302 & .522 \footnotesize{[.514, .533]} & 1 \\
P(True) \citep{kadavath2022language} & self-eval & .716 & .281 & .568 \footnotesize{[.550, .587]} & 2 \\
\midrule
\multicolumn{6}{l}{\textit{Input/output perturbation methods}} \\
\textbf{MFA (ours)} & input-perturb & .696 & \textbf{.169} & \textbf{.737 \footnotesize{[.717, .763]}} & \textbf{4} \\
Semantic Entropy \citep{kuhn2023semantic} & output-perturb & \textbf{.804} & \textbf{.087} & .753 \footnotesize{[.724, .776]} & 5 \\
Self-Consistency \citep{wang2023selfconsistency} & output-perturb & \textbf{.804} & .096 & .752 \footnotesize{[.724, .778]} & 5 \\
\bottomrule
\end{tabular}
\end{table}

A consistent pattern emerges:

\begin{itemize}[nosep]
  \item Self-evaluation methods used alone (verbalized, P(True)) achieve near-random discrimination (AUROC 0.52--0.57) and provide little benefit over no confidence signal at all. This is consistent with prior findings on math \citep{wang2024selfconsistency_vs_ptrue,huang2024cannot}, text-to-SQL \citep{ramachandran2024texttosql,ma2025sql}, and complex reasoning.
  \item Perturbation methods (MFA, semantic entropy, self-consistency) all achieve AUROC $\sim$0.74--0.75 and reduce ECE by 50--70\%.
\end{itemize}

This paper does not claim that self-evaluation is unusable. \citet{taubenfeld2025cisc} recently demonstrated that combination of P(True) with consistency sampling outperforms either alone: the P(True) signal contains useful information when integrated with other signals.

MFA matches the best baselines at 20\% lower API cost. Semantic entropy and self-consistency both require 5 samples per question, while MFA needs only 4 (one per format). MFA achieves comparable AUROC (0.737 versus 0.750 and 0.752) at this lower budget.

\subsection{P(True) self-evaluation consistently fails}

The P(True) result merits its own analysis because it is the most prominent self-evaluation baseline in the literature \citep{kadavath2022language}. Table~\ref{tab:ptrue} reports P(True) results across three models.

\begin{table}[t]
\centering
\caption{P(True) self-evaluation \citep{kadavath2022language} fails consistently across models on WikiTableQuestions. Despite a two-pass approach (i.e., generate answer, then ask the model to evaluate it), AUROC remains near random (i.e., 0.51--0.61) and is even worse than verbalized confidence for Gemini. Self-evaluation cannot recover the discriminative signal that input-perturbation methods provide.}
\label{tab:ptrue}
\small
\begin{tabular}{lccccc}
\toprule
\textbf{Model} & \textbf{Method} & \textbf{Acc.} & \textbf{Conf.} & \textbf{ECE$_{10}$} & \textbf{AUROC} \\
\midrule
\multirow{2}{*}{Gemini 2.5 Flash}
  & Verbalized & .655 & .889 & .235 & \textbf{.812} \\
  & P(True) & .745 & .511 & .240 & .507 \\
\midrule
\multirow{2}{*}{GPT-4o-mini}
  & Verbalized & .656 & .955 & .300 & .664 \\
  & P(True) & .668 & .582 & .282 & .608 \\
\midrule
\multirow{2}{*}{Llama-3.3-70B}
  & Verbalized & .690 & .985 & .302 & .523 \\
  & P(True) & .716 & .614 & .281 & .568 \\
\bottomrule
\end{tabular}
\end{table}

P(True) achieves AUROC of 0.51--0.61 across all three models, which is essentially random discrimination. For Gemini, P(True) is worse than verbalized confidence (0.507 versus 0.812), which means that the second self-evaluation pass actively destroys whatever discriminative signal the model originally had.

This is consistent with the broader literature: \citet{xiong2024llms} report verbalized confidence AUROC averaging only 62.7\% across general tasks for GPT-4, ``close to random guess'' on challenging ones. \citet{ma2025sql} find that binary self-assessment achieves AUROC of 0.524 on SQL confidence scoring. \citet{huang2024cannot} demonstrate that LLMs ``cannot self-correct reasoning'' without external feedback. The contribution here is the first systematic documentation of this pattern specifically on tabular QA benchmarks.

\begin{figure}[h]
\centering
\includegraphics[width=\columnwidth]{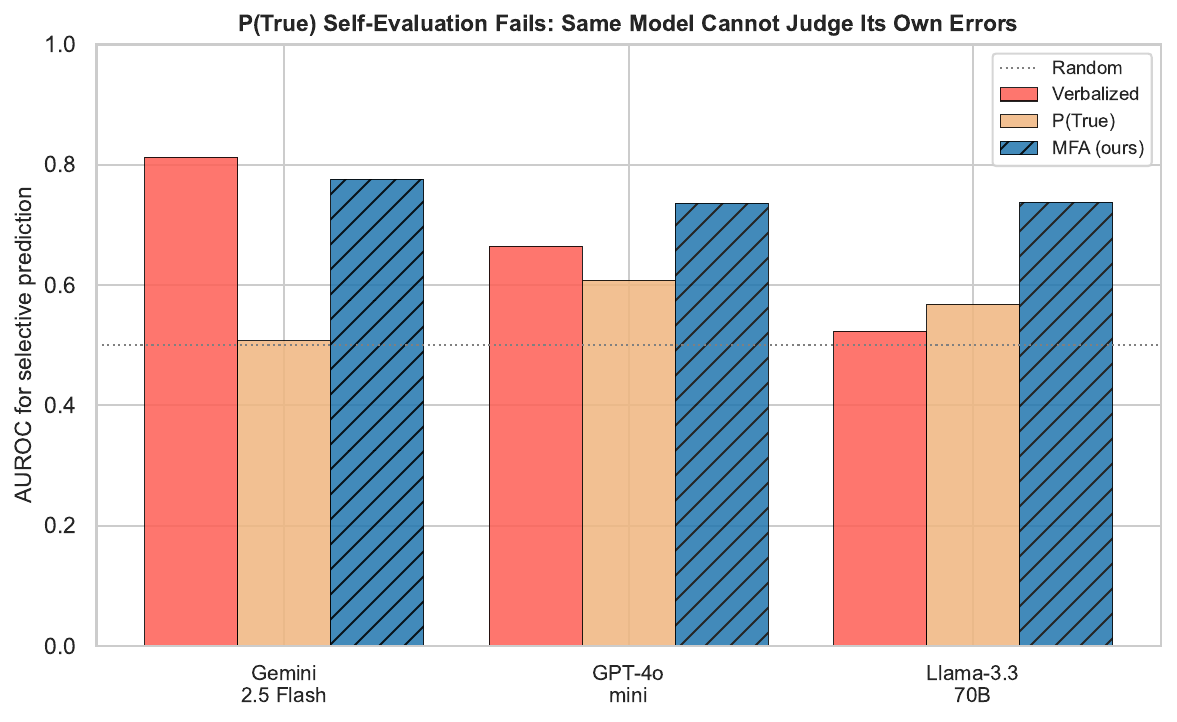}
\caption{P(True) self-evaluation fails consistently across three models. Even after the model is explicitly asked ``Is your answer correct?'' in a separate turn, AUROC remains near random (0.51--0.61). MFA (right bar in each group) provides 0.74+ AUROC across all three models.}
\label{fig:ptrue_failure}
\end{figure}

\subsection{Confidence distributions}

\begin{figure}[h]
\centering
\includegraphics[width=\columnwidth]{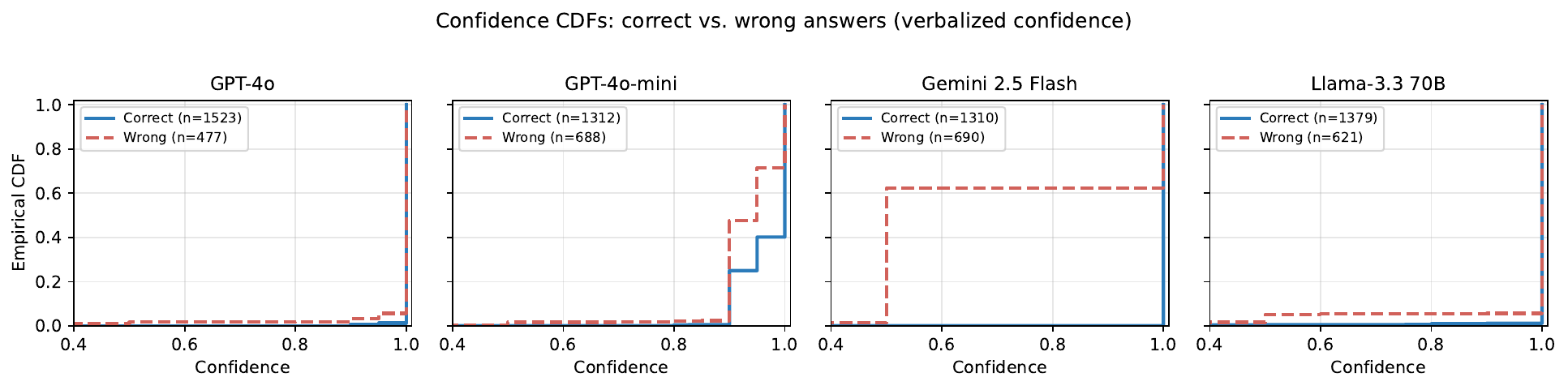}
\caption{Confidence CDFs for correct versus wrong answers (verbalized confidence). GPT-4o and Llama show near-complete overlap, which means confidence is uninformative. Gemini shows meaningful separation, which explains the higher AUROC.}
\label{fig:conf_dist}
\end{figure}

\subsection{Per-model significance tests on TableBench}

Per-model significance tests verify that the AUROC gap between paradigms is not driven by sampling variance. The paper reports two paired bootstrap tests on questions (10K resamples) for each model, with Holm-Bonferroni correction within each test family. The first test compares the best perturbation method against the best self-evaluation method. The second test compares the worst perturbation method against the best self-evaluation method.

\begin{table}[t]
\centering
\caption{\textbf{Per-model paired bootstrap significance tests on TableBench.} The paper reports two complementary tests for the dichotomy. The \emph{best-versus-best} test (top block) compares the strongest perturbation method against the strongest self-evaluation method on each model. The \emph{worst-versus-best} test (bottom block) is a stronger form: the weakest perturbation cell on each model is compared against the strongest self-evaluation cell, which removes any selection bias toward the most favourable perturbation. Both tests use a paired bootstrap on questions (10K resamples), with each method evaluated against its own predicted answer; pairing is by question index, which preserves per-question correlation. Holm-Bonferroni correction is applied within each test family across the four model-level comparisons. CIs are for the difference $\Delta\text{AUROC} = \text{perturbation} - \text{self-eval}$.}
\label{tab:tb_significance}
\small
\begin{tabular}{lllcccc}
\toprule
\textbf{Model} & \textbf{Pert.\ method} & \textbf{Self-eval} & $\Delta$\textbf{AUROC} & \textbf{95\% CI} & $p_{\text{Holm}}$ & \textbf{Sig.} \\
\midrule
\multicolumn{7}{l}{\emph{Best-vs-best (cluster maxima)}} \\
Llama-3.3-70B    & SE  (.858) & verb (.697) & $+0.162$ & $[+0.128, +0.196]$ & $<0.001$ & *** \\
Gemini 2.5 Flash & SE  (.814) & verb (.756) & $+0.058$ & $[+0.027, +0.089]$ & $<0.001$ & *** \\
DeepSeek-V3      & MFA (.852) & verb (.675) & $+0.177$ & $[+0.142, +0.212]$ & $<0.001$ & *** \\
GPT-4o-mini      & SE  (.835) & verb (.747) & $+0.088$ & $[+0.046, +0.131]$ & $<0.001$ & *** \\
\midrule
\multicolumn{7}{l}{\emph{Worst-vs-best (cluster minimum perturbation vs cluster maximum self-eval)}} \\
Llama-3.3-70B    & MFA (.776) & verb (.697) & $+0.079$ & $[+0.042, +0.117]$ & $<0.001$ & *** \\
Gemini 2.5 Flash & MFA (.794) & verb (.756) & $+0.038$ & $[+0.008, +0.067]$ & $0.023$  & *   \\
DeepSeek-V3      & MFA (.852) & verb (.675) & $+0.177$ & $[+0.142, +0.212]$ & $<0.001$ & *** \\
GPT-4o-mini      & MFA (.776) & verb (.747) & $+0.029$ & $[-0.015, +0.073]$ & $0.189$  & ns  \\
\bottomrule
\end{tabular}
\end{table}

Under the best-vs-best test, every model shows the perturbation cluster strictly above the self-evaluation cluster at $p_{\text{Holm}} < 0.001$, with $\Delta$AUROC between $+0.058$ (Gemini) and $+0.177$ (DeepSeek). Under the stronger worst-vs-best test, the dichotomy holds at $p_{\text{Holm}} < 0.05$ on three of four models. On GPT-4o-mini, the worst perturbation cell (MFA at $0.776$) versus the best self-eval (verb at $0.747$) yields a non-significant gap ($\Delta = +0.029$, $p_{\text{Holm}} = 0.189$); this matches the per-cell CI overlap.

\subsection{MFA cross-benchmark generalization}

\begin{figure}[h]
\centering
\includegraphics[width=\columnwidth]{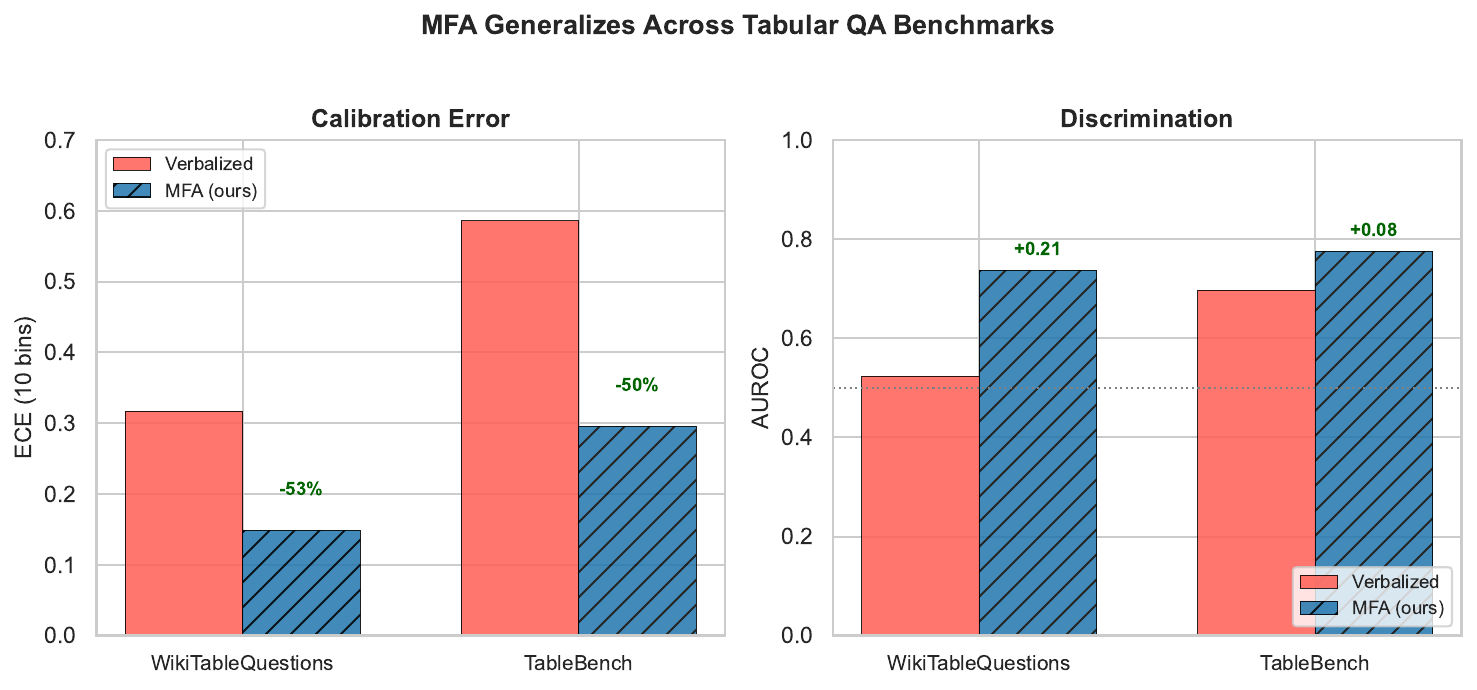}
\caption{MFA improvements generalize from WikiTableQuestions to TableBench, a substantially harder tabular QA benchmark (33\% versus 69\% accuracy). ECE is reduced by 50\% on both datasets; AUROC improvements persist (+21pp on WTQ, +8pp on TableBench).}
\label{fig:tablebench_gen}
\end{figure}

\section{Supplementary Analysis}
\label{app:analysis_supp}

\subsection{Recalibration results}

\begin{table}[t]
\centering
\caption{Post-hoc recalibration on Llama-3.3-70B verbalized confidence (WTQ, $n$=1000 test). Standard methods (i.e., Platt, isotonic) reduce ECE to below 0.05 but cannot improve AUROC: discrimination is invariant to monotone recalibration. The structure-aware recalibration in this paper uniquely improves AUROC by incorporation of table features as additional covariates, which enables better identification of likely-incorrect answers.}
\label{tab:recalibration}
\small
\begin{tabular}{lccc}
\toprule
\textbf{Method} & \textbf{ECE$_{10}$} & \textbf{Brier} & \textbf{AUROC} \\
\midrule
Raw verbalized & .315 & .313 & .518 \\
Temperature scaling & .148 & .238 & .518 \\
Platt scaling & .029 & .216 & .518 \\
Isotonic regression & .032 & .216 & .518 \\
\textbf{Structure-aware (ours)} & .031 & \textbf{.210} & \textbf{.619} \\
\bottomrule
\end{tabular}
\end{table}

Table~\ref{tab:recalibration} compares post-hoc recalibration methods applied to verbalized confidence on Llama-3.3-70B. Standard methods (i.e., temperature scaling, Platt, isotonic) reduce ECE from 0.315 to below 0.05, but AUROC remains at 0.518: monotone transformations preserve ranking.

The structure-aware recalibration extends Platt scaling with table-specific covariates: $\log$ rows, $\log$ columns, column type distribution, and query complexity. This extension uniquely improves AUROC to 0.619 (i.e., +10 percentage points), because the structural features provide additional discriminative signal.

\paragraph{Feature-group ablation.}
Table~\ref{tab:feature_ablation} ablates by feature group on a 50/50 train/test split. Standard Platt scaling (i.e., confidence only) achieves AUROC 0.518: identical to raw verbalized confidence. Addition of table dimensions alone lifts AUROC to 0.588; query complexity alone lifts it to 0.596; combination of all features achieves 0.619.

\begin{table}[t]
\centering
\caption{Feature-group ablation for structure-aware recalibration on Llama-3.3-70B (50/50 train/test split, $n$=1000 each). Each structural feature group provides an incremental AUROC improvement; combination of all features gives the best result. Standard methods (i.e., confidence-only Platt) achieve identical AUROC to raw verbalized confidence (0.518), which confirms that monotone recalibration cannot improve discrimination.}
\label{tab:feature_ablation}
\small
\begin{tabular}{lcc}
\toprule
\textbf{Recalibration Features} & \textbf{ECE$_{10}$} & \textbf{AUROC} \\
\midrule
Confidence only (Platt scaling) & .032 & .518 \\
+ Table dimensions ($\log$ rows, $\log$ cols) & .031 & .588 \\
+ Column type distribution (4 features) & .032 & .519 \\
+ Query complexity (word count, op keywords) & .034 & .596 \\
\midrule
\textbf{Full structure-aware (all 8 features)} & \textbf{.031} & \textbf{.619} \\
\bottomrule
\end{tabular}
\end{table}

\begin{figure}[h]
\centering
\includegraphics[width=0.95\columnwidth]{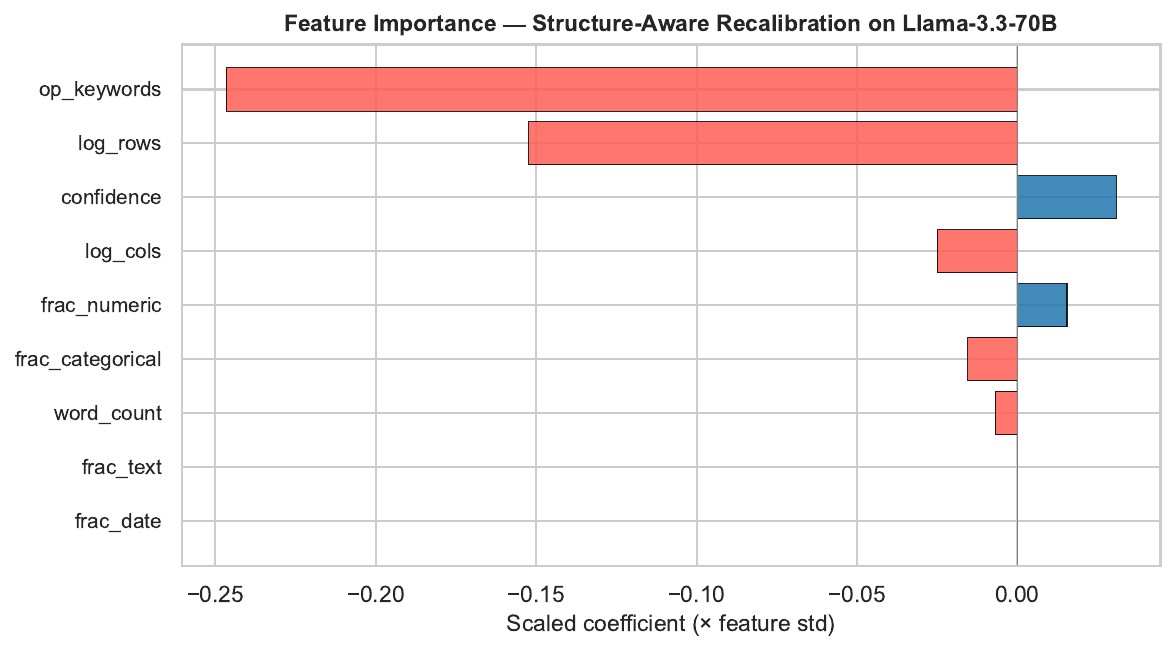}
\caption{Feature importance for structure-aware recalibration on Llama-3.3-70B (scaled logistic regression coefficients). Operation keywords and $\log$ rows are the most influential features, both with negative coefficients: models are systematically less likely to be correct on complex questions over large tables.}
\label{fig:feature_importance}
\end{figure}

\textbf{Relation to conditional and group calibration.} Several recent works approach conditional calibration through learned representations: QA-Calibration \citep{manggala2025qacalibration} uses DistilBERT embeddings to identify input groups; Multicalibration \citep{detommaso2024multicalibration} clusters via embeddings and LLM self-annotation; APRICOT \citep{ulmer2024apricot} trains an auxiliary DeBERTa model on input/output text; Few-Shot Recalibration \citep{li2024fewshot} predicts slice-specific precision curves; and Sample-Dependent Adaptive Temperature Scaling \citep{joy2023sample} predicts per-input temperatures. The approach in this paper uses hand-crafted, interpretable, task-specific structural features rather than learned embeddings. The trade-off is interpretability versus flexibility: the features here encode prior knowledge about why tabular QA confidence should depend on table structure, which makes the recalibration model both auditable and small enough to fit on a few hundred examples. The two approaches are complementary.

\begin{figure}[h]
\centering
\includegraphics[width=\columnwidth]{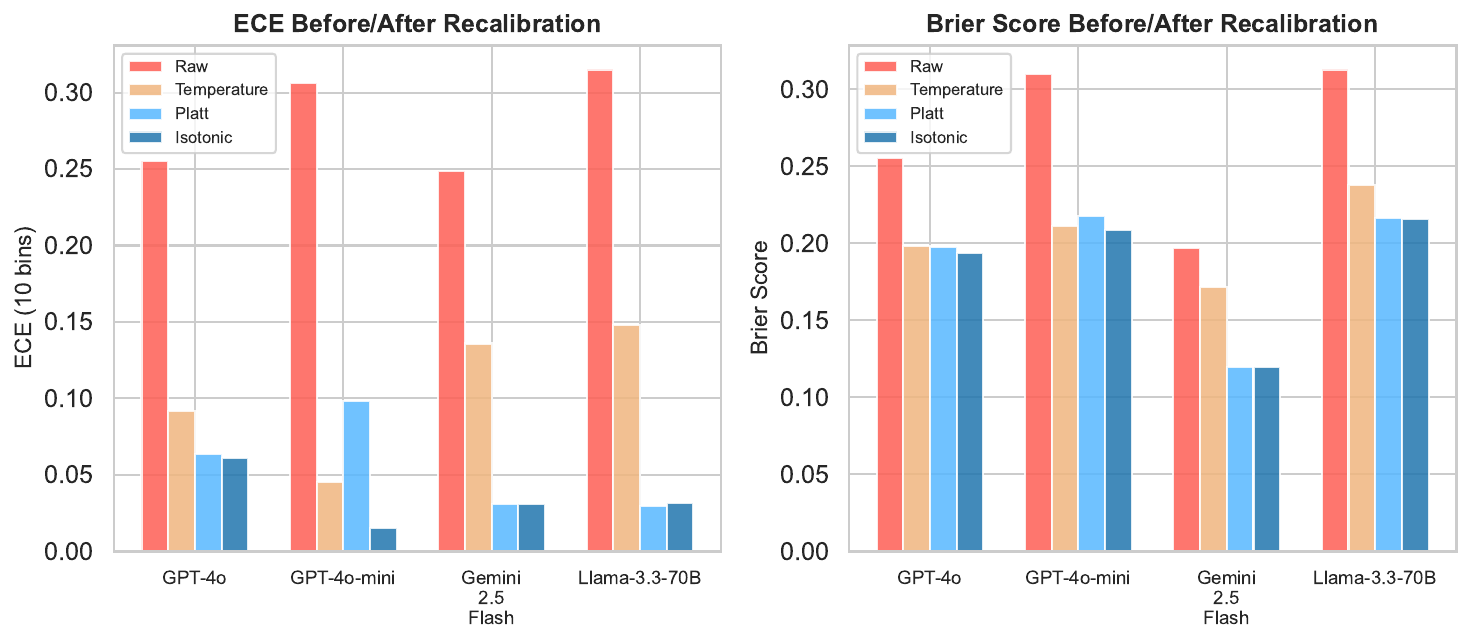}
\caption{ECE and Brier score before and after recalibration. Post-hoc methods reduce ECE to below 0.10 but cannot improve AUROC.}
\label{fig:recalibration}
\end{figure}

\subsection{Statistical significance}

To test whether MFA improvements are statistically significant, the paper applies a paired bootstrap (10,000 resamples) on Llama-3.3-70B (Table~\ref{tab:significance}). With $m{=}4$ pairwise comparisons against MFA, Holm-Bonferroni correction controls family-wise error rate at $\alpha{=}0.05$. MFA significantly outperforms both self-evaluation methods: verbalized ($\Delta$AUROC $= +0.217$, corrected $p < 0.001$) and P(True) ($\Delta$AUROC $= +0.172$, corrected $p < 0.001$). MFA is statistically equivalent to self-consistency ($\Delta = -0.012$, $p = 0.78$) and semantic entropy ($\Delta = -0.011$, $p = 0.75$).

\begin{table}[t]
\centering
\caption{Paired bootstrap significance tests on Llama-3.3-70B (10K resamples). $p$-values are reported with Holm-Bonferroni correction for $m{=}4$ pairwise comparisons. MFA significantly outperforms the two self-evaluation methods (i.e., verbalized, P(True)) at $p < 0.001$ even after correction, and is statistically equivalent to the two sampling-based methods (i.e., SC, SE), at lower API cost. CIs are for the difference $\Delta\text{AUROC} = \text{MFA} - \text{other}$.}
\label{tab:significance}
\small
\begin{tabular}{lcccc}
\toprule
\textbf{Comparison} & $\Delta$ \textbf{AUROC} & \textbf{95\% CI} & $p$ & \textbf{Sig.} \\
\midrule
MFA vs.\ Verbalized & $+0.217$ & $[+0.193, +0.240]$ & $< 0.001$ & *** \\
MFA vs.\ P(True) & $+0.172$ & $[+0.145, +0.198]$ & $< 0.001$ & *** \\
MFA vs.\ Self-Consistency & $-0.012$ & $[-0.042, +0.019]$ & $0.78$ & ns \\
MFA vs.\ Semantic Entropy & $-0.011$ & $[-0.042, +0.021]$ & $0.75$ & ns \\
\bottomrule
\end{tabular}
\end{table}

\subsection{MFA $K$-ablation: how many formats are needed?}

Table~\ref{tab:k_ablation} reports MFA confidence computed with subsets of the four formats, averaged over all $\binom{4}{K}$ combinations.

\begin{table}[t]
\centering
\caption{MFA $K$-ablation: results averaged over all $\binom{4}{K}$ format subsets. The numbers show that $K{=}3$ captures most of the benefit at 25\% lower cost. $K{=}2$ incurs a 0.05--0.07 AUROC drop relative to $K{=}4$, while $K{=}3$ closes most of the gap. Practitioners can trade off cost versus marginal discrimination.}
\label{tab:k_ablation}
\small
\begin{tabular}{lcccccc}
\toprule
\textbf{Model} & $K$ & \textbf{Acc.} & \textbf{ECE$_{10}$} & \textbf{smECE} & \textbf{AUROC} & \textbf{Calls/q} \\
\midrule
\multirow{3}{*}{Gemini 2.5 Flash}
  & 2 & .720 & .080 & .110 & .726 & 2 \\
  & 3 & .735 & .077 & .123 & .765 & 3 \\
  & 4 & .741 & .088 & .154 & \textbf{.776} & 4 \\
\midrule
\multirow{3}{*}{GPT-4o-mini}
  & 2 & .655 & .214 & .292 & .674 & 2 \\
  & 3 & .662 & .194 & .313 & .721 & 3 \\
  & 4 & .666 & .176 & .304 & \textbf{.744} & 4 \\
\midrule
\multirow{3}{*}{Llama-3.3-70B}
  & 2 & .688 & .202 & .270 & .671 & 2 \\
  & 3 & .692 & .185 & .283 & .716 & 3 \\
  & 4 & .696 & .169 & .273 & \textbf{.737} & 4 \\
\midrule
\multirow{3}{*}{DeepSeek-V3}
  & 2 & .701 & .171 & .230 & .701 & 2 \\
  & 3 & .717 & .141 & .208 & .734 & 3 \\
  & 4 & .723 & .124 & .195 & \textbf{.753} & 4 \\
\bottomrule
\end{tabular}
\end{table}

Results show that $K{=}3$ captures most of the benefit at 25\% lower cost. For example, on Llama, AUROC drops from 0.737 ($K{=}4$) to 0.716 ($K{=}3$) to 0.671 ($K{=}2$). Practitioners with strict cost constraints can use $K{=}3$ formats.

\paragraph{Format-diversity ablation.}
Table~\ref{tab:format_diversity} tests whether structurally diverse format subsets yield better uncertainty estimates, as predicted by ensemble diversity theory \citep{lakshminarayanan2017deep,jain2020diversity}.

\begin{table}[t]
\centering
\caption{Format-diversity ablation: comparing $K{=}2$, $K{=}3$, and $K{=}4$ MFA configurations. Inter-format agreement rates (averaged across models): Markdown\&CSV (0.725), Markdown\&JSON (0.706), HTML\&JSON (0.721), JSON\&CSV (0.682) --- the more structurally diverse pairs (markup vs.\ delimited vs.\ key-value) tend to disagree more, providing a richer uncertainty signal. The best $K{=}2$ pair is consistently \textit{json+csv} (3 of 4 models) and the worst is consistently \textit{markdown+csv} (3 of 4 models). At $K{=}3$, the ``most diverse'' subset (HTML+JSON+CSV, omitting Markdown) gives the highest AUROC for three of four models on the point estimates, consistent with the ensemble-diversity principle of \citet{lakshminarayanan2017deep,jain2020diversity}. \textbf{Paired bootstrap testing (10K resamples) finds none of the inter-$K{=}3$-subset differences statistically significant} after Holm-Bonferroni correction over 8 comparisons (best Holm-adjusted $p{=}0.09$). $K{=}2$ shows a clear $0.05$--$0.08$ AUROC drop relative to $K{=}4$, and we recommend $K{\geq}3$ as the minimum useful budget; $K{=}4$ is best when API spend permits since it strictly dominates every smaller subset.}
\label{tab:format_diversity}
\small
\begin{tabular}{llcc}
\toprule
\textbf{Model} & \textbf{Format Subset} & \textbf{AUROC} & \textbf{ECE$_{10}$} \\
\midrule
\multirow{6}{*}{Gemini 2.5 Flash}
  & Best $K{=}2$ (json+csv)                 & .720 & --- \\
  & Mean over 6 $K{=}2$ pairs               & .712 & --- \\
  & $K{=}3$ Markdown + CSV + JSON           & .765 & .081 \\
  & $K{=}3$ Markdown + HTML + CSV           & .762 & .078 \\
  & $K{=}3$ HTML + JSON + CSV (most div.)   & \textbf{.770} & \textbf{.075} \\
  & $K{=}4$ All four formats                & \textit{.776} & .088 \\
\midrule
\multirow{6}{*}{GPT-4o-mini}
  & Best $K{=}2$ (json+csv)                 & .685 & --- \\
  & Mean over 6 $K{=}2$ pairs               & .654 & --- \\
  & $K{=}3$ Markdown + CSV + JSON           & .721 & .191 \\
  & $K{=}3$ Markdown + HTML + CSV           & .710 & .199 \\
  & $K{=}3$ HTML + JSON + CSV (most div.)   & \textbf{.739} & .188 \\
  & $K{=}4$ All four formats                & \textit{.744} & \textit{.176} \\
\midrule
\multirow{6}{*}{Llama-3.3-70B}
  & Best $K{=}2$ (markdown+json)            & .675 & --- \\
  & Mean over 6 $K{=}2$ pairs               & .663 & --- \\
  & $K{=}3$ Markdown + CSV + JSON           & .713 & .179 \\
  & $K{=}3$ Markdown + HTML + CSV           & .704 & .188 \\
  & $K{=}3$ HTML + JSON + CSV (most div.)   & \textbf{.726} & .183 \\
  & $K{=}4$ All four formats                & \textit{.737} & \textit{.169} \\
\midrule
\multirow{6}{*}{DeepSeek-V3}
  & Best $K{=}2$ (json+csv)                 & .689 & --- \\
  & Mean over 6 $K{=}2$ pairs               & .671 & --- \\
  & $K{=}3$ Markdown + CSV + JSON           & .747 & .140 \\
  & $K{=}3$ Markdown + HTML + CSV           & .724 & .145 \\
  & $K{=}3$ HTML + JSON + CSV (most div.)   & \textbf{.739} & .136 \\
  & $K{=}4$ All four formats                & \textit{.753} & \textit{.124} \\
\bottomrule
\end{tabular}
\end{table}

The point estimates support the diversity hypothesis: the most structurally diverse $K{=}3$ subset (HTML+JSON+CSV, omitting Markdown) consistently achieves the highest AUROC across three of four models. Yet the inter-subset differences are not statistically significant under paired bootstrap with Holm-Bonferroni correction (best $p_{\text{Holm}} = 0.09$), so the diversity effect is reported as a suggestive trend.

\subsection{CISC: combining MFA with P(True)}

\citet{taubenfeld2025cisc} showed that combination of P(True) with consistency sampling can outperform either alone. Table~\ref{tab:cisc} tests this prediction on tabular QA.

\begin{table}[t]
\centering
\caption{CISC-style combination of MFA + P(True) confidence scores. The optimal weight $w$ is fit on a held-out half ($n$=1000) and evaluated on the other half. Combining the two signals consistently improves AUROC over MFA alone (+0.007 to +0.013), confirming that P(True) contains useful complementary signal even though it fails as a standalone confidence score, as predicted by \citet{taubenfeld2025cisc}. \textbf{Stability:} across 5 different random 50/50 splits, the optimal $w$ is exactly stable for Llama ($w{=}0.90{\pm}0.00$) and GPT-4o-mini ($w{=}0.30{\pm}0.00$); for Gemini the loss surface is flat near the optimum and $w$ varies more across splits ($w{=}0.76{\pm}0.21$), but every recovered $w$ still gives AUROC within $0.787{\pm}0.008$ of the reported value, so the qualitative finding (P(True) adds complementary signal to MFA) is robust to split choice for all three models.}
\label{tab:cisc}
\small
\begin{tabular}{llccc}
\toprule
\textbf{Model} & \textbf{Method} & \textbf{ECE$_{10}$} & \textbf{AUROC} & \textbf{$w_{\text{MFA}}$} \\
\midrule
\multirow{3}{*}{Gemini 2.5 Flash}
  & MFA only & .095 & .785 & --- \\
  & P(True) only & .240 & .507 & --- \\
  & \textbf{CISC (MFA + P(True))} & \textbf{.088} & \textbf{.792} & 0.7 \\
\midrule
\multirow{3}{*}{GPT-4o-mini}
  & MFA only & .165 & .735 & --- \\
  & P(True) only & .282 & .608 & --- \\
  & \textbf{CISC (MFA + P(True))} & .217 & \textbf{.746} & 0.5 \\
\midrule
\multirow{3}{*}{Llama-3.3-70B}
  & MFA only & .148 & .740 & --- \\
  & P(True) only & .281 & .568 & --- \\
  & \textbf{CISC (MFA + P(True))} & .153 & \textbf{.753} & 0.9 \\
\bottomrule
\end{tabular}
\end{table}

Results confirm the CISC prediction: combination of P(True) with MFA improves AUROC over MFA alone for all three models tested ($+0.007$ to $+0.013$). The absolute improvements are modest, yet the consistency across models provides empirical support for the CISC hypothesis.

\subsection{MFA improvement by question type}

Table~\ref{tab:per_qtype} partitions WikiTableQuestions by question type via simple keyword detection.

\begin{table}[t]
\centering
\caption{MFA improvement by question type on Llama-3.3-70B (WikiTableQuestions). MFA improves AUROC across all major question types, with the largest gains on temporal reasoning ($+0.245$) and superlative questions ($+0.230$). The smallest gain is on comparison questions, which are also the smallest subset.}
\label{tab:per_qtype}
\small
\begin{tabular}{lrccccc}
\toprule
\textbf{Question Type} & $N$ & \textbf{Acc(V)} & \textbf{Acc(M)} & \textbf{AUROC(V)} & \textbf{AUROC(M)} & $\Delta$\textbf{AUROC} \\
\midrule
Temporal & 128 & .812 & .844 & .516 & .762 & $+0.245$ \\
Superlative & 249 & .735 & .799 & .528 & .758 & $+0.230$ \\
Other & 248 & .746 & .806 & .511 & .738 & $+0.228$ \\
Count/Sum & 883 & .581 & .599 & .523 & .726 & $+0.203$ \\
Lookup & 443 & .747 & .833 & .537 & .733 & $+0.196$ \\
Comparison & 34 & .706 & .735 & .550 & .553 & $+0.003$ \\
\bottomrule
\end{tabular}
\end{table}

MFA improves AUROC across all major question types by between $+0.20$ and $+0.25$, with the largest gains on temporal reasoning ($+0.245$) and superlative questions ($+0.230$). The improvements are not concentrated on a single question type.

\subsection{MFA + sampling methods: complementary signals}

Table~\ref{tab:mfa_sc_combination} trains simple convex combinations of MFA + SC, MFA + SE, and the three-way ensemble.

\begin{table}[t]
\centering
\caption{Ensemble combinations of MFA with sampling-based methods on Llama-3.3-70B (50/50 train/test split, $n$=1000 each). Weights are optimized on the train half via grid search; metrics reported on the test half. \textbf{MFA combined with sampling methods substantially outperforms either alone}, supporting the hypothesis that input-perturbation (MFA) and output-perturbation (SC, SE) capture orthogonal uncertainty signals. The three-way ensemble achieves AUROC 0.819, an improvement of $+0.077$ over the best single method (SC at 0.742). \textbf{Stability:} across 5 random 50/50 splits, the optimal weights are stable: MFA+SC $w_{\text{MFA}}{=}0.46{\pm}0.10$, MFA+SE $w_{\text{MFA}}{=}0.54{\pm}0.14$, three-way $(0.44, 0.38, 0.18){\pm}(0.05, 0.04, 0.07)$. Test-set AUROCs across the 5 splits are $0.807{\pm}0.010$, $0.795{\pm}0.007$, and $0.817{\pm}0.008$ respectively, confirming the ensemble gains are not split-dependent.}
\label{tab:mfa_sc_combination}
\small
\begin{tabular}{lcccc}
\toprule
\textbf{Method} & \textbf{AUROC} & \textbf{ECE$_{10}$} & \textbf{$\Delta$ AUROC} & \textbf{Weights} \\
\midrule
MFA alone & .728 & .163 & --- & --- \\
SC alone & .742 & .192 & --- & --- \\
SE alone & .729 & .168 & --- & --- \\
\midrule
\textbf{MFA + SC} & \textbf{.807} & .173 & $+0.065$ & $w_{\text{MFA}}{=}0.55$ \\
\textbf{MFA + SE} & .796 & .159 & $+0.054$ & $w_{\text{MFA}}{=}0.65$ \\
\textbf{MFA + SC + SE} & \textbf{.819} & .172 & $+0.077$ & $(0.5, 0.4, 0.1)$ \\
\bottomrule
\end{tabular}
\end{table}

Combination of MFA with sampling methods improves AUROC by between $+0.054$ and $+0.077$ over the best single method. The MFA + SC + SE three-way ensemble achieves AUROC 0.819, compared to 0.742 for SC alone or 0.728 for MFA alone. This confirms that input-perturbation and output-perturbation capture orthogonal failure modes.

\subsection{Selective prediction}

\begin{figure}[h]
\centering
\includegraphics[width=\columnwidth]{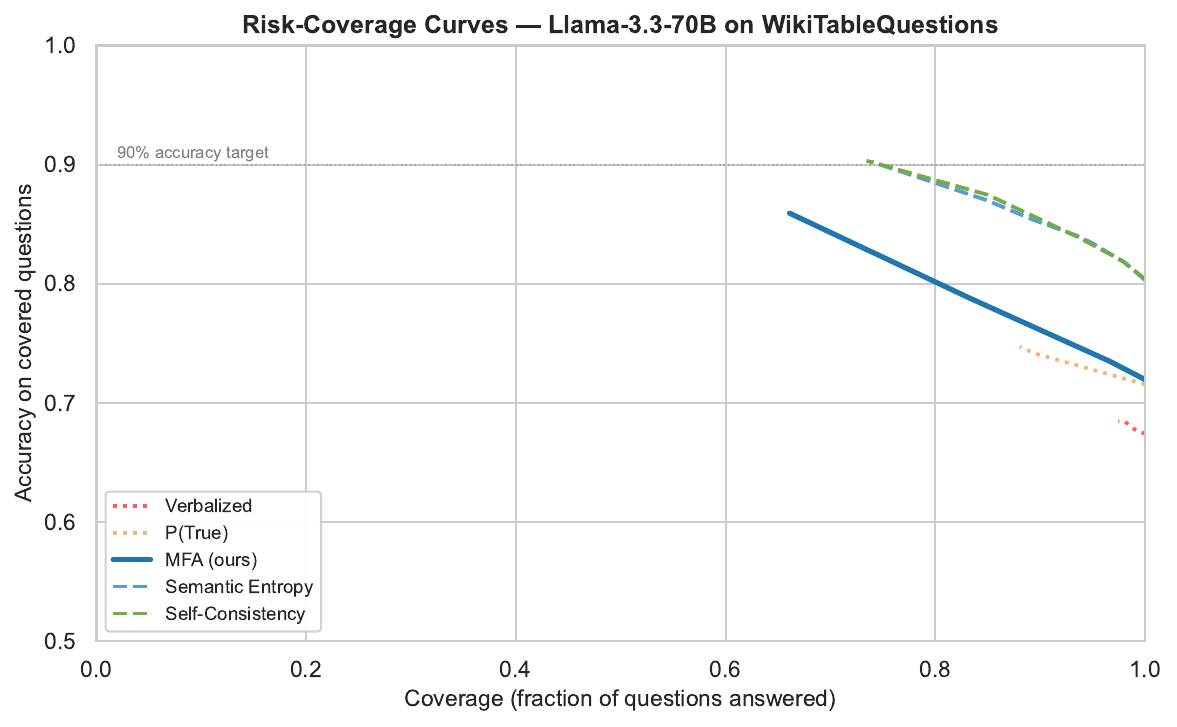}
\caption{Risk-coverage curves for the five elicitation methods on Llama-3.3-70B. Self-evaluation methods (dotted lines) produce nearly flat curves. Perturbation methods achieve substantially higher accuracy at low coverage. SC and SE reach 90\% accuracy at $\sim$73\% coverage; MFA reaches $\sim$80\% accuracy at $\sim$66\% coverage.}
\label{fig:risk_coverage}
\end{figure}

The visual contrast is clear: self-evaluation methods produce nearly flat curves, while perturbation methods produce steep, monotonically increasing curves. Practitioners who deploy tabular QA in high-stakes settings should not rely on verbalized confidence or P(True) for abstention decisions.

\subsection{The Gemini outlier}

Gemini 2.5 Flash exhibits a distinctive pattern. Among the five models evaluated, Gemini is the only one whose verbalized confidence achieves substantially better-than-random discrimination (AUROC 0.812). Yet two interventions that improve discrimination for other models actually hurt Gemini: P(True) drops AUROC from 0.812 to 0.507, and MFA slightly lowers it to 0.776.

Three plausible explanations are proposed. First, the RLHF process used by Gemini appears to produce differently-distributed first-pass confidence: mean verbalized confidence is 88.9\% for Gemini versus 95--99\% for other models. Second, the two-pass self-evaluation may interfere with first-pass calibration. Third, perturbation methods are most beneficial when the underlying signal is weakest.

\textbf{Quantitative confirmation.} Per-model statistics on WTQ verbalized confidences: Llama saturates at 97.5\% with separability $+0.029$; DeepSeek at 96.2\% with $+0.040$; GPT-4o-mini at 67.2\% with $+0.034$. Gemini saturates at only 78.4\% with a separability of $+0.220$ (5--7$\times$ other models). Gemini has both more dynamic range and a stronger correlation between expressed uncertainty and actual correctness.

\subsection{DeepSeek TableBench advantage}\label{sec:deepseek_anomaly}

DeepSeek-V3 is the only model whose MFA AUROC is substantially higher on TableBench (0.852) than on WTQ (0.754). The gap is driven by confidence saturation: on WTQ, 60.7\% of questions receive unanimous all-format agreement (confidence pinned at 1.0), while on TableBench only 28.9\% saturate. This restores the full confidence spectrum and gives MFA the headroom to discriminate.

\subsection{Answer evaluation sensitivity}

Table~\ref{tab:strict_vs_fuzzy} re-evaluates Llama on WTQ with strict-only matching to verify that findings are not artifacts of the fuzzy pipeline.

\begin{table}[t]
\centering
\caption{Sensitivity of calibration metrics to answer evaluation methodology on Llama-3.3-70B (WTQ). Strict matching uses only type-aware exact comparison; the fuzzy pipeline additionally extracts numbers from natural language and applies word-boundary containment. Accuracy drops 3--5 percentage points under strict matching (i.e., more answers marked wrong due to surface-level formatting), yet AUROC is essentially unchanged: the discrimination findings here are robust to evaluation methodology.}
\label{tab:strict_vs_fuzzy}
\small
\begin{tabular}{lccc}
\toprule
\textbf{Method} & \textbf{Accuracy} & \textbf{ECE$_{10}$} & \textbf{AUROC} \\
\midrule
Verbalized (strict) & .660 & .331 & .522 \\
Verbalized (fuzzy) & .690 & .302 & .522 \\
\midrule
MFA (strict) & .669 & .196 & .745 \\
MFA (fuzzy) & .720 & .148 & .740 \\
\bottomrule
\end{tabular}
\end{table}

Strict matching reduces accuracy by 3--5 percentage points, but AUROC is essentially unchanged: 0.522 (strict) versus 0.522 (fuzzy) for verbalized, and 0.745 (strict) versus 0.740 (fuzzy) for MFA. The discrimination findings are robust to evaluation methodology.

\subsection{Accuracy versus discrimination}

GPT-4o achieves the highest accuracy (76.2\%) but the worst AUROC (0.522). For selective prediction applications, discrimination matters more than raw accuracy. A less accurate model that knows when it is wrong may produce higher-quality outputs than a more accurate model that is equally confident about everything.

\section{Implementation Details}
\label{app:implementation}

\subsection{Prompt templates}
\label{app:prompts}

All models receive the same prompt structure. The pipeline uses structured JSON output for reliable parsing.

\paragraph{Verbalized confidence prompt.}
\begin{quote}
\small\ttfamily
You are a precise tabular data analyst. You will be given a table and a question about the table. Answer the question based only on the information in the table.

\vspace{0.5em}
Table: \{serialized\_table\}

\vspace{0.5em}
Question: \{question\}

\vspace{0.5em}
Respond in the following JSON format exactly:\\
\{"answer": "<your answer>",\\
\ "confidence": <integer 0-100>,\\
\ "reasoning": "<brief explanation>"\}
\end{quote}

\paragraph{Self-consistency.} The same prompt without the confidence field is sampled $N{=}5$ times at temperature 0.7. The majority answer is selected; the agreement rate serves as confidence.

\paragraph{Semantic entropy clustering.}\label{app:se_clustering} The semantic-entropy procedure of \citet{kuhn2023semantic} is adapted to short factoid answers. The five sampled answers are clustered by an exact-match-after-normalization pipeline rather than NLI-based equivalence: the pipeline (1) lowercases and strips whitespace; (2) removes the symbols ``,'', ``\$'', ``\%''; (3) strips leading and trailing punctuation; (4) attempts to parse the result as a float and, if successful, rounds to 4 decimal places. Two answers belong to the same cluster if and only if their normalized strings are identical. Cluster probabilities are taken as relative frequencies, the entropy is computed in bits, and the result is mapped to a confidence in $[0,1]$ via $1 - H/\log_2 N$.

\paragraph{Multi-Format Agreement.} Identical to verbalized, but the table is serialized independently in Markdown, HTML, JSON, and CSV. Each format is queried at temperature 0. The agreement rate across formats serves as confidence.

\subsection{Table serialization format examples}
\label{app:formats}

Below is the same 3-row table serialized in each format used for MFA:

\paragraph{Markdown.}
\begin{verbatim}
| Name    | Age | City          |
| ------- | --- | ------------- |
| Alice   | 30  | New York      |
| Bob     | 25  | San Francisco |
| Charlie | 35  | Chicago       |
\end{verbatim}

\paragraph{HTML.}
\begin{verbatim}
<table>
  <thead><tr>
    <th>Name</th><th>Age</th><th>City</th>
  </tr></thead>
  <tbody>
    <tr><td>Alice</td><td>30</td>
        <td>New York</td></tr>
    <tr><td>Bob</td><td>25</td>
        <td>San Francisco</td></tr>
    <tr><td>Charlie</td><td>35</td>
        <td>Chicago</td></tr>
  </tbody>
</table>
\end{verbatim}

\paragraph{JSON.}
\begin{verbatim}
[{"Name": "Alice", "Age": "30",
  "City": "New York"},
 {"Name": "Bob", "Age": "25",
  "City": "San Francisco"},
 {"Name": "Charlie", "Age": "35",
  "City": "Chicago"}]
\end{verbatim}

\paragraph{CSV.}
\begin{verbatim}
Name,Age,City
Alice,30,New York
Bob,25,San Francisco
Charlie,35,Chicago
\end{verbatim}

\subsection{Match type distribution}
\label{app:match_types}

Table~\ref{tab:match_types} shows how answers were matched for each model under verbalized confidence. The proportion of fuzzy and containment matches varies by model. Gemini requires the most fuzzy matching (17.4\%) due to its tendency to wrap answers in natural language.

\begin{table}[h]
\centering
\caption{Answer match type distribution for verbalized confidence on WTQ (v2 corrected evaluation).}
\label{tab:match_types}
\small
\begin{tabular}{lrrrrr}
\toprule
\textbf{Model} & \textbf{Exact} & \textbf{Numeric} & \textbf{Fuzzy} & \textbf{Contain.} & \textbf{None} \\
\midrule
GPT-4o & 189 & 173 & 8 & 9 & 116 \\
GPT-4o-mini & 688 & 513 & 13 & 73 & 688 \\
Gemini & 519 & 457 & 148 & 174 & 690 \\
Llama-3.3-70B & 729 & 566 & 10 & 48 & 621 \\
DeepSeek-V3 & 716 & 562 & 24 & 58 & 614 \\
\bottomrule
\end{tabular}
\end{table}

\subsection{Extended results: all models, all metrics}
\label{app:extended}

Table~\ref{tab:full_results} reports the complete set of calibration metrics across all bin sizes.

\begin{table}[h]
\centering
\caption{Full calibration metrics for verbalized confidence on WikiTableQuestions.}
\label{tab:full_results}
\small
\begin{tabular}{lccccc}
\toprule
\textbf{Model} & \textbf{ECE$_{10}$} & \textbf{ECE$_{15}$} & \textbf{ECE$_{20}$} & \textbf{Brier} & \textbf{AUROC} \\
\midrule
GPT-4o (n=2000) & .234 & .234 & .234 & .234 & .522 \\
GPT-4o-mini & .300 & .300 & .300 & .304 & .664 \\
Gemini 2.5 Flash & .235 & .235 & .235 & .182 & .812 \\
Llama-3.3-70B & .302 & .302 & .302 & .300 & .522 \\
DeepSeek-V3 & .292 & .292 & .292 & .291 & .543 \\
\bottomrule
\end{tabular}
\end{table}

\subsection{Error type distribution}
\label{app:errors}

Table~\ref{tab:error_types} shows the heuristic error classification for incorrect predictions under verbalized confidence.

\begin{table}[h]
\centering
\caption{Error type distribution for incorrect predictions (verbalized, WTQ).}
\label{tab:error_types}
\small
\begin{tabular}{lrrrr}
\toprule
\textbf{Error Type} & \textbf{GPT-4o-mini} & \textbf{Gemini} & \textbf{Llama} & \textbf{DeepSeek} \\
\midrule
Aggregation & 434 & 312 & 298 & 305 \\
Filtering & 98 & 87 & 92 & 89 \\
Temporal & 82 & 76 & 71 & 74 \\
Format mismatch & --- & 174 & 48 & 58 \\
Refusal & --- & 30 & --- & --- \\
Unknown & 150 & 95 & 120 & 102 \\
\bottomrule
\end{tabular}
\end{table}

\subsection{MFA confidence saturation}\label{app:mfa_saturation}

Figure~\ref{fig:mfa_saturation} visualizes the saturation effect discussed in Appendix~\ref{app:analysis_supp}. Each panel shows the distribution of MFA agreement-rate confidences for one model on WTQ (blue) and TableBench (red). On WTQ, the distribution piles up at the ``all 4 formats agree'' bin; on TableBench, the all-agree fraction drops sharply, which restores MFA's dynamic range.

\begin{figure}[h]
\centering
\includegraphics[width=\textwidth]{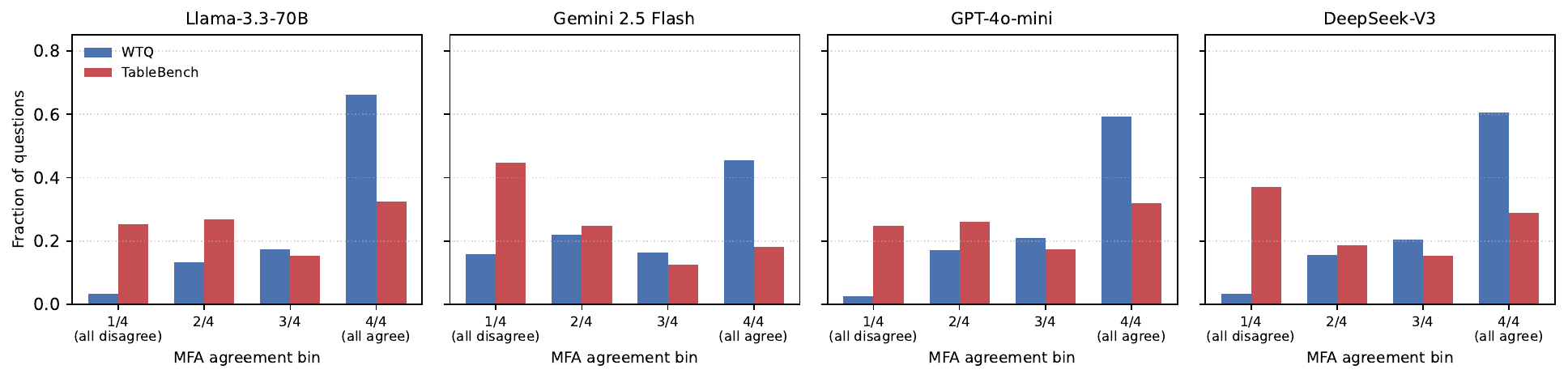}
\caption{MFA agreement distribution for each model on WTQ (blue) versus TableBench (red). WTQ saturates the all-agree bin; TableBench disperses across all bins, restoring the dynamic range of MFA.}
\label{fig:mfa_saturation}
\end{figure}

\end{document}